\documentclass[review]{elsarticle}

\newcommand{\citeauthoryearkt}[1]{\citeauthor{#1}~\citeyear{#1}}

\usepackage[margin=2cm]{geometry}
\usepackage{amssymb}
\usepackage{amsmath}
\usepackage{graphicx}
\usepackage{gensymb}
\usepackage{mathrsfs}
\usepackage{xcolor}
\usepackage{float}    
\usepackage{booktabs}
\usepackage{caption}
\usepackage{makecell}
\usepackage{subcaption}
\usepackage{color}
\usepackage[normalem]{ulem} 
\usepackage{cancel}
\usepackage{setspace}
\usepackage{lineno}

\definecolor{ktred}{rgb}{0.90, 0.10, 0.1}
\definecolor{ktblack}{rgb}{0.0, 0.0, 0.0}
\definecolor{ktgreen}{rgb}{0.0, 0.50, 0.0}
\definecolor{nwblue}{rgb}{0.0, 0.0, 0.6}

\newcommand{\ktaddrem}[1]{\textcolor{ktblack}{#1}}

\newcommand{\kteqrem}[1]{\ifmmode\text{\color{ktred}\sout{\ensuremath{#1}}}\else\sout{#1}\fi}

\journal{Journal of Engineering Mechanics}

\begin{document}

\begin{frontmatter}
\title{A numerical study into neural network surrogate model performance for uncertainty propagation}
\author[add1]{Noah Wade\corref{cor1}}
\ead{noah.s.wade2.ctr@us.navy.mil}
\author[add2]{Kirubel Teferra} 
\cortext[cor1]{Corresponding author}

\affiliation[add1]{organization={ASEE Postdoctoral Associate},
            addressline={U.S. Naval Research Laboratory},
            city={Washington},
            state={DC},
            postcode={20375},
            country={USA}}

\affiliation[add2]{organization={U.S. Naval Research Laboratory},
            city={Washington},
            postcode={20375},
            state={DC},
            country={USA}}
\begin{abstract}
	Neural network surrogate models have emerged as a promising approach to model solution fields for a wide variety of boundary value problems encountered in physical modeling. Stochastic problems represent an area of particularly high interest because of the potential to significantly reduce, or completely circumvent, the repeated evaluation of expensive forward models via traditional numerical solvers when conducting parametric analysis. However, \ktaddrem{many studies found in the literature} primarily focus on the ability of \ktaddrem{neural network surrogate models} to represent deterministic samples or mean field solutions and largely overlook \ktaddrem{surrogate model performance at the tails of the distribution. The present study examines in detail the ability of neural network surrogate models to capture the full distribution of solution fields over the entire probability space, while emphasis is placed at the tails of the distribution}. Serving as a canonical problem is the heat conduction equation with a highly stochastic source term, inducing extremely large variation in the thermal solution field. Comparisons are made between a classic feed-forward fully connected network and a Deep Operator Network architecture, using both data-driven and physics-informed loss functions. Results show that the worst-case prediction errors of the neural networks are an order of magnitude larger than the mean field error, highlighting the importance of the outlier samples. The large errors associated with extreme samples result from the networks having to extrapolate beyond the bounds of the training data. \ktaddrem{A method for identifying these samples is presented along with a discussion of potential approaches to account of their errors.} Among the models considered, the fully connected neural network trained using a weak form residual loss performs best in handling these extrapolated inputs, achieving the highest prediction accuracy for the numerically produced datasets. 
\end{abstract}

\begin{keyword}
\ktaddrem{Neural network surrogate models \sep Physics informed neural networks \sep Deep operator networks \sep Uncertainty propagation}
\end{keyword}

\end{frontmatter}

\section{Introduction}
Neural networks are increasingly being used as surrogate models across a wide range of applications within computational mechanics. Traditionally, neural networks are trained using large amounts of labeled training data, which for many mechanics applications is either computationally expensive to obtain or incomplete due to the limited availability of experimental data. To address this, a subclass of networks called Physics Informed Neural Networks (PINNs), introduced in \citeauthoryearkt{Lagaris_1998} and popularized in \citeauthoryearkt{RAISSI_2019}, has been developed to leverage fundamental physical laws within the training process and alleviate the need for big data. By incorporating the residual of the governing differential equations into the training loss function, several studies have successfully trained neural networks to solve thermal (\citeauthoryearkt{CAI_2021_physics}, \citeauthoryearkt{Sukumar_2022}), fluid flow (\citeauthoryearkt{XU_2021_WEAK}, \citeauthoryearkt{Cai_2021_review}), fracture mechanics (\citeauthoryearkt{GOSWAMI_2022}), and multifidelity problems (\citeauthoryearkt{Howard_2023}, \citeauthoryearkt{Li_2020}), among other problems in computational mechanics. 

However, the PINN framework still faces many limitations, including difficulty in training, lack of generalization, and high computational cost (\citeauthoryearkt{Zhang_2024}, \citeauthoryearkt{Markidis_2021}, \citeauthoryearkt{Mata_2023}), making it challenging to identify applications where they clearly outperform existing numerical methods (\citeauthoryearkt{Grossmann_2024}, \citeauthoryearkt{Sacchetti_2022}).  Many PINNs also suffer from slow training convergence, a challenge that becomes more severe as the input sample space increases in dimension and variability. Furthermore, adaptive training methods developed to improve computational efficiency in single-instantiation problems, such as adaptive collocation placement (\citeauthoryearkt{Hou_2023}), importance sampling  (\citeauthoryearkt{Nabian_2021}, \citeauthoryearkt{ Daw_2022}), and adaptive weight rebalancing (\citeauthoryearkt{Xiang_2022}), are likely to struggle to achieve similar performance gains for stochastic problems as the deterministic ones because of the competing contributions among training samples.

While the problems discussed above develop neural network surrogate models for a single sample instantiation of a set of parameters (i.e., deterministic problems), their true promise lies in their potential to serve as low cost response surfaces for solutions fields over high-dimensional sample spaces. This capability is critical for stochastic problems where traditional methods require repeated evaluation of expensive numerical solvers to propagate uncertainty. Leveraging PINNs for such tasks could enable rapid evaluation across the parameter space, facilitating uncertainty propagation and quantification at a fraction of the computational cost. The aim of this study is to \ktaddrem{examine the capability of neural network surrogate models} to represent stochastic boundary value problems (SBVP) in such cases.

There have been a number of studies that utilize the PINN framework to include an uncertain parameter as an input, often with minimal distinction from the spatial and temporal coordinate inputs in terms of network architecture (\citeauthoryearkt{Zhang_2022}, \citeauthoryearkt{Yang_2019}, \citeauthoryearkt{Lutjens_2021}, \citeauthoryearkt{Hu_2024}).  It is worth noting that many example problems in the literature treat a partial differential equation (PDE) coefficient (e.g. thermal conductivity) as the uncertain parameter; however, the solution field variance tends to be insensitive to PDE coefficients in comparison to source terms or boundary conditions. For example, \citeauthoryearkt{XU_2021_WEAK} include a small variability in the hydraulic conductivity to show that PINNs can handle stochastic coefficients but the focus of the study is on the required number of collocation points and their placement rather than the stochastic variability introduced by the hydraulic conductivity, which is fairly insignificant. Other works have successfully integrated uncertainty propagation with PINNs, such as \citeauthoryearkt{Zhang_2020}, who use model space learning for time-dependent stochastic PDEs to enable uncertainty propagation, \citeauthoryearkt{Lin_2024}, who incorporates variability via a Monte Carlo PINN approach for the Boltzmann transport equation, and \citeauthoryearkt{Nabian_2019} who use PINNs to predict the mean and standard deviation of the response field for diffusion and heat conduction problems. However, each of these studies primarily focus on lower order statistics of the solution fields and not on evaluating neural network models' ability in capturing rare events. 

\ktaddrem{Another approach to consider for propagating input random fields is to use neural operators, which seek to learn a mapping from input functions (e.g., infinite dimensional field variables that arise as coefficients, forcing terms, or boundary conditions in the governing PDEs) to solution field variables rather than learning finite dimensional input-output relationships. This results in a change in the neural network surrogate architecture that better reflects the input-output data relationship than the classical feed forward architecture. Two notable examples of neural operator architectures are the Fourier Neural Operator (FNO) (\citeauthoryearkt{Li_2020}) and the Deep Operator Network (DeepONet) (\citeauthoryearkt{Lu_2021}). These network structures have been used alongside physics informed learning to advance the performance of neural network surrogate models. Some examples include: \citeauthoryearkt{Li_2024}, who combine coarse-resolution training data with physics informed loss constraints within the FNO framework to accurately learn solution operators for parametric PDEs, and \citeauthoryearkt{Lu_2021} who use DeepONets to decouple the learning of an input function from that of the spatial coordinates in order to decrease the generalization error in comparison to a fully connected neural network. There have also been several works which have used neural operators to solve stochastic problems, such as \citeauthoryearkt{Zhang_2022} who apply DeepONets to stochastic differential equations to predict the mean and variance of a stochastic Poisson equation using supervised learning, and \citeauthoryearkt{Kontolati_2024} who develop a latent space input to a DeepONet for time-dependent PDEs. Some additional examples of neural network surrogate models for uncertainty quantification examples can be found in the following references (\citeauthoryearkt{exenberger2025deep}, \citeauthoryearkt{Kumar_2025}, \citeauthoryearkt{Karumuri_2025}, \citeauthoryearkt{Kong_2020}, \citeauthoryearkt{Pickering_2022}).}

\ktaddrem{The main challenge with using neural network surrogate models to solve SBVPs relates to neural networks ability excel in interpolation and conversely fail in extrapolation. This results in diminished accuracy in the probability tails due to the scarcity of training samples in these regions, which is itself a consequence of their low probability of occurrence. Maintaining prediction accuracy far from the mean is therefore non-trivial unless training datasets and methods are specifically designed to address the distribution tails (\citeauthoryearkt{Wabartha_2021}). This challenge is well studied in the field of image classification, and is denoted as the long tail problem (\citeauthoryearkt{Anderson_2006,Cui_2019}). \citeauthoryearkt{Yang_2022} presents a review of ten methods (along with numerous variations) for training image classifiers to better classify tail class images (i.e. those images which are rare and only have a small number of samples in the training set). Among these, cost-sensitive weighting, where additional weight factors are assigned to training samples based on their rarity, and data augmentation, where the training data is augmented with additional samples from the tail classes, show high levels of success. For regression problems, addressing the long tail problem through weight modification is less developed, although some studies have been conducted. For example, \citeauthoryearkt{Rudy_2023} assess the performance of different weighting strategies that apply greater weights to the tails through estimates of the probability density function of the output and compare against the standard mean square error. They observe that an improved performance is reached at the tails in comparison to mean square error, yet there remain significant errors across the full sample space. It is worth noting that this study is conducted on single instantiations of chaotic systems rather than an ensemble of solution fields, and thus, it is questionable if the same conclusions apply across samples in a SBVP problem.} 

\ktaddrem{An alternative perspective is to enrich training data sets with additional samples near the region of interest in the probability space. For example, \citeauthoryearkt{Song_2024} utilizes importance sampling to enhance the reliability of predictions of a neural network surrogate model for the limit state function for a reliability problem involving penetration of chloride ions in concrete. They accomplish this by first conducting an initial training of the network to ascertain the relationship between the input space of random variables and the problem's limit state boundary. Then, using this information, they incorporate importance sampling to retrain the network, focusing its training on those samples used to predict that limit state. A similar concept is employed in \citeauthoryearkt{Pickering_2022} that utilizes the concept of an acquisition function for selecting extreme samples to be included in training data sets. While these approaches enable training surrogate models targeted specifically for regions in the sample space with low probability they often simultaneously sacrifice accuracy in the so-called body of the sample space (i.e., regions with higher probability).}

\ktaddrem{The perspective taken in this study is to evaluate neural network surrogate model performance across the entire sample space.} To gain insight into the relative performance of the neural network surrogate models described above, this study presents the results of a comparative analysis on a problem characterized by extremely high stochasticity. Specifically, it examines the heat conduction equation with a source field exhibiting very large variance. The network architectures evaluated include feed-forward neural networks and DeepONets, while the training loss functions compared are a data-only loss and a PINN-based loss incorporating the PDE residual. The models are primarily ranked based on their prediction accuracy in the tails of the thermal field distribution, as results show that errors for the most extreme samples are approximately ten times larger than the mean error. Additionally, this study investigates the use of instance-based adaptive weighting to guide training towards improved accuracy for these extreme cases. \ktaddrem{Finally, an analysis is provided which shows that the most extreme errors are correlated with outliers to the training data, and a method for identifying them is presented. This knowledge can help to identify when using a surrogate model is appropriate and how to augment training data to account for regions with larger errors. Thus, the primary contribution of this work is the analysis, and corresponding discussion, of a numerical investigation which extensively evaluates the performance of ubiquitously used neural network architectures and loss functions for uncertainty propagation under highly variable solution fields.} Overall, the findings provide insight into the convergence and accuracy of different architectures, loss function types, and training techniques in low-probability regions.

The paper is outlined as follows. Section \ref{Methods} describes the formulation of the stochastic PDE used in this study (Section \ref{Source_generation}) and includes descriptions of the various neural network architectures (Section \ref{NN_Architectures}), loss functions (Section \ref{NN_Loss}), and evaluation metrics (Section \ref{Evaluation_Metrics}). Section \ref{Results} provides the results and performance evaluations of the surrogate models for a series of numerical studies, followed by concluding remarks in Section \ref{Conclusions}.

\section{Methodology} \label{Methods}
This study evaluates the performance of various neural network surrogate models and associated training techniques for accurately predicting the response surface of random fields that exhibit large spatial and magnitude variations in the context of uncertainty propagation. The models' performance are evaluated via a SBVP with a highly variable source field, described in Section \ref{Source_generation}. The three neural network surrogate models considered are:

\begin{enumerate}
	\item \textbf{Data-Driven Network (DDN)}: A fully connected neural network trained purely on labeled data (finite element solutions). This model serves as a benchmark for comparison, allowing an evaluation of whether the physics-informed approaches can outperform traditional data-driven methods.
	
	\item \textbf{Physics-Informed Neural Network (PINN)}: A model that uses the same architecture as the DDN but incorporates a physics-informed loss function derived from the governing equations. This model tests the effectiveness of embedding physics into the learning process.
	
	\item \textbf{Deep Operator Network (DeepONet)}: A physics-informed model that uses a DeepONet architecture. This model features the same loss functions as the PINN model but tests a different network architecture which may be better suited for the problem at hand.
\end{enumerate}

Each model will be tasked with the nonlinear mapping of the stochastic parameters $\boldsymbol\xi \in \mathbb{R}^{N_d}$, which define the source field, to the solution field variable, \ktaddrem{$T(\mathbf{x});\  \mathbf{x} \in \mathbb{R}^d$}, expressed as:

\begin{equation} \label{Mapping}
	\ktaddrem{\mathcal{G}: \boldsymbol \xi \mapsto T(\mathbf{x})}
\end{equation}

\noindent \ktaddrem{with $N_d$ and $d$ being the stochastic dimension and physical dimension of the problem, respectively.} The models are evaluated primarily based on their accuracy in predicting the quantity of interest over a standardized set of test data. Although computational performance and efficiency are relevant considerations, they are treated as secondary objectives in this study. Instead, the focus is placed on analyzing the magnitude and distribution of prediction errors across individual test samples. This emphasis is motivated by the nature of stochastic problems, which often produce outliers. As will be shown, these outliers can exhibit significantly larger errors compared to samples within one or two standard deviations of the mean, and accounting for them is critical if these approaches are to be used for reliability problems. 

The remainder of Section \ref{Methods} is dedicated to defining the problem set up, outlining the network architectures and loss functions, and defining the error metrics used for performance evaluation.

\subsection{Problem Set Up} \label{Source_generation}
A steady-state heat conduction problem with stochastic source field is chosen for this study due to its sensitivity to small variations in the source field. This sensitivity enables the construction of a highly stochastic source field using only a few random variables, resulting in wide ranging solution response fields, which presents a significant challenge for accurate neural network surrogate model prediction.

The governing equation for steady-state heat conduction, subject only to essential boundary conditions, is given by:

\begin{equation} \label{PDE}
	\begin{split}
		\ktaddrem{-k \cdot \nabla^2 T = S} \\
		T(\Gamma) = T_o
	\end{split}
\end{equation}

\noindent where $k$ is the thermal conductivity, $T$ is the temperature, $S$ defines the thermal source, and all points on the boundary, represented by $\Gamma$, are set to a fixed temperature, $T_o$. The corresponding weak form of Eq. \ref{PDE} is:

\begin{equation} \label{weak_long}
	\int_\Omega (\nabla w)^T \cdot k \nabla T \; d\Omega - \int_\Omega w S \; d\Omega = 0 \quad \forall w
\end{equation}

\noindent where $w$ is an arbitrary valid test function. Restricting the problem to a two-dimensional domain, \ktaddrem{$\mathbf{x}=\left(x,y\right)$}, the weak form can be further simplified to:

\begin{equation} \label{weak}
	\begin{split}
		\ktaddrem{\int_\Omega  k \left [ \frac{\partial w}{\partial x}  \frac{\partial T(x,y)}{\partial x}  + \frac{\partial w}{\partial y}  \frac{\partial T(x,y)}{\partial y} \right ] d\Omega \ - \int_\Omega wS(x,y)  \; d\Omega  = 0 \quad \forall w}
	\end{split} 
\end{equation}

The source term is derived from an underlying random field $S'(x,y)$, defined with respect to its  Karhunen--Lo\`eve (K-L) decomposition:

\begin{equation} \label{S}
	S'(x, y) \approx \sum_{i=1}^{N_d} \sqrt{\lambda_i} \, v_i(x, y) \, \xi_i
\end{equation}

\noindent where $N_d$ is the chosen number of uniform random variables of $\boldsymbol \xi$ (four in this study), and the eigenvalues \( \lambda_i \) and eigenfunctions \( v_i(x,y) \) are derived from an arbitrarily chosen autocorrelation function, \( R \), defined as:

\begin{equation} \label{Auto Correlation}
	\begin{aligned}
		R(\tau_x,\tau_y) = e^{-0.03 \cdot (\tau_x + \tau_y)^2} + e^{-0.06 \cdot (\tau_x + \tau_y)^3}
	\end{aligned}
\end{equation}

\noindent and \( \tau_x \) and \( \tau_y \) denote the spatial lag.

Then a local, nonlinear transformation of the underlying random field $S'(x,y)$ is applied to give the source term random field \( S(x,y) \) as:

\begin{equation} \label{Scaled S}
	S = \alpha [A S'^2 + B S'^4 + C S'^6]
\end{equation}

Defining $S(x,y)$ in this way enables the generation of a distribution of highly skewed thermal fields, which can be scaled and distorted by adjusting a small set of arbitrary parameters, [$\alpha, A, B, C$].

\begin{figure}[ht]
	\includegraphics[width=1\textwidth]{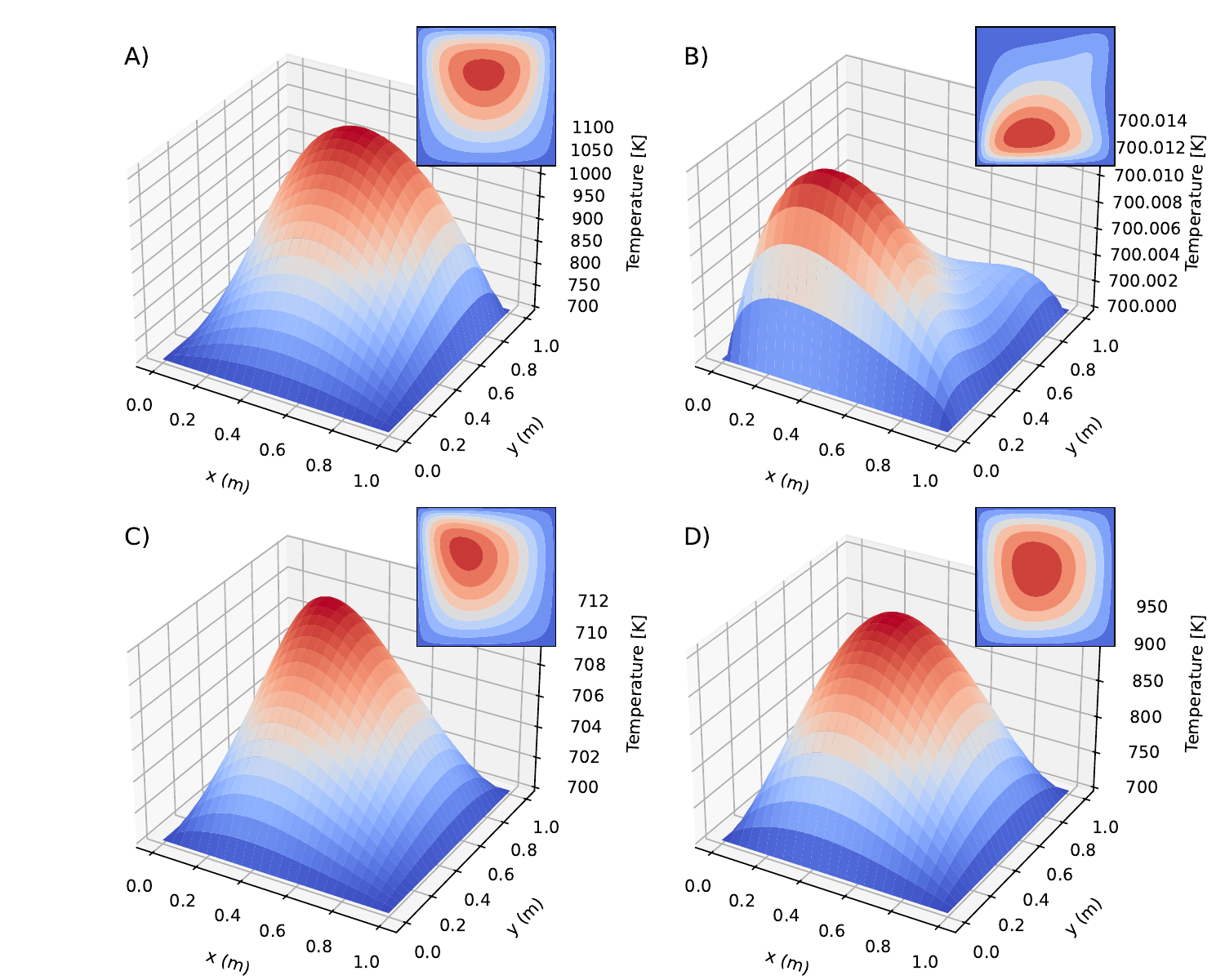}
	\caption{Solution fields from Source 2 (defined in Table \ref{Distributions_Statistics}) showing the: A) Maximum Peak, B) Minimum Peak, C) Median Temperature Peak, and  D) Random Peak, from a dataset of 2k random samples.}
	\label{4_solutions}
\end{figure}

A few of the thermal solution fields $T(\mathbf{x})$ generated in this manner are shown in Figure \ref{4_solutions}. The source terms for each field are generated by drawing samples for the first four eigenmodes of $S'(x,y)$, where the parameters $\xi_i$ for $i = 1, \dots, 4$ are treated as independent, identically distributed uniform random variables in the range $[-3, 3]$. Following the transformation of Eq. \eqref{Scaled S}, each realization of $S(x,y)$ is passed through a finite element solver to compute the corresponding steady-state temperature field, which serves as the ground truth. \ktaddrem{The numerical accuracy is assed through a mesh convergence study where the mesh is successively doubled in each direction and the finite element solution field for randomly selected samples of the source field is recomputed. The difference between a [50$\times$50] and [100$\times$100] mesh is deemed negligible both quantitatively and qualitatively. The average relative $\ell^2$-norm error of over 100 samples between a [50$\times$50] and [100$\times$100] mesh is 1.6e$^{-3}$. Qualitatively, plots of randomly selected section cuts of the solution field for the two meshes overlayed on each other have nearly imperceptible differences. Thus, [50$\times$50] mesh size is considered a reasonable compromise considering the computationally intensiveness of this study. It is further noted that by maintaining consistency in the mesh size among the finite element ground truth and the neural network surrogate models, the finite element errors, although present, play a diminished role in the comparison among the models.}

\begin{figure}[ht]
	\centering
	\includegraphics[width=1\textwidth]{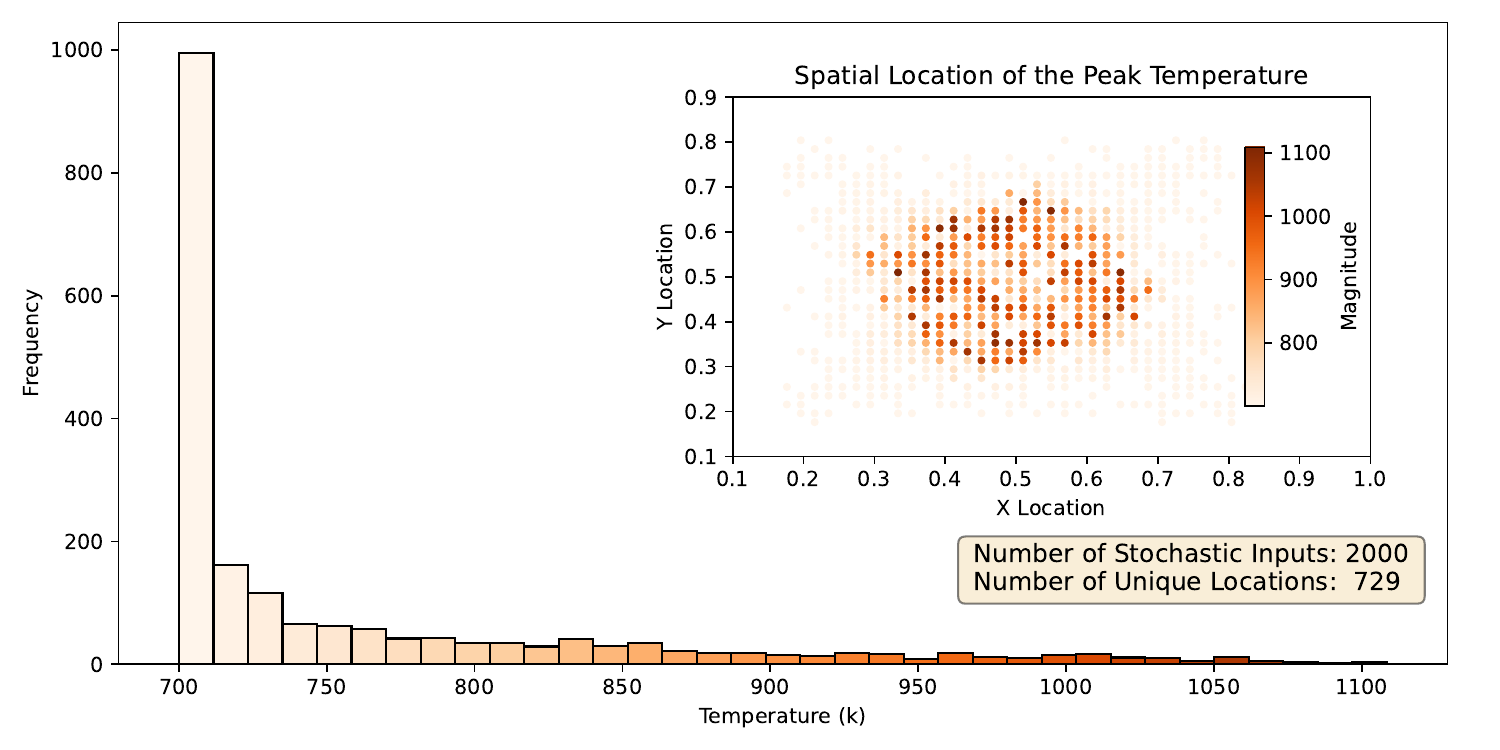}
	\caption{Histogram of the maximum temperatures of the solution field over 2000 samples for Source 2. The range covers over 400 $K$ with most of the extreme peaks being located near the center of the domain. The number of unique maximum temperature locations is determined by counting the finite element nodes that serve as the maximum temperature point for at least one sample.}
	\label{Extreme_dist_hist}
\end{figure}

Two datasets, defined as Source 1 and Source 2, of 2000 independent random samples are generated using parameter values given in Table \ref{Distributions_Statistics}. \ktaddrem{The first set of data, Source 1, features sources which have a small maximum temperature variation of about 80 $K$ across the domain. The second set of data, Source 2, has a much larger temperature range of just over 400 $K$. By fixing the Dirichlet boundary conditions as constant for both cases, the solution field due to Source 2 has larger and more concentrated peaks, usually near the center of the domain, while the solution field for Source 1 has its larger values more smoothly distributed through the interior of the domain.} A key feature of each of the datasets is that the distribution of the maximum of the solution field temperatures are heavy tailed. The histogram of the maximum temperature of each sample for Source 2 is shown in Figure \ref{Extreme_dist_hist} along with a visualization of the approximate locations of each temperature peak. Due to this heavy tail, the behavior of this distribution poses a significant challenge to predict because the maximum temperatures are so far removed from the mean.

\begin{table}[ht]
	\centering
	\caption{Statistics for the magnitude of peak temperature of 2000 random samples across two random sources. The parameter values refer to Eq. \eqref{Scaled S}.}
	\label{Distributions_Statistics}
	\begin{tabular}{lcc}
		\toprule
		& \makecell{\textbf{Random}\\\textbf{Source 1}} & \makecell{\textbf{Random}\\\textbf{Source 2}} \\
		\midrule
		\textbf{Parameter Values}               & [2, 2, 4, 6] & [10, 10, 20, 30] \\
		\textbf{Mean Peak Temperature}               & 712.8 $K$ & 758.2 $K$ \\
		\textbf{Median Peak Temperature}             & 702.9 $K$ & 711.9 $K$ \\
		\textbf{Max Peak Temperature}                & 783.4 $K$& 1108.6 $K$ \\
		\bottomrule
	\end{tabular}
\end{table}

\subsection{Network Architectures} \label{NN_Architectures}

\begin{figure}[ht]
	\centering
	\begin{subfigure}[t]{0.48\textwidth}
		\centering
		\includegraphics[width=\textwidth]{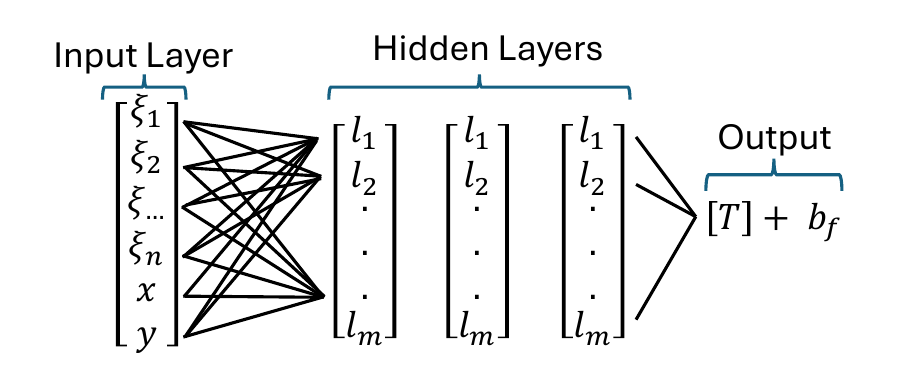}
		\caption{Fully Connected Neural Network Structure}
		\label{FNN_structure}
	\end{subfigure}
	\hfill
	\begin{subfigure}[t]{0.48\textwidth}
		\centering
		\includegraphics[width=\textwidth]{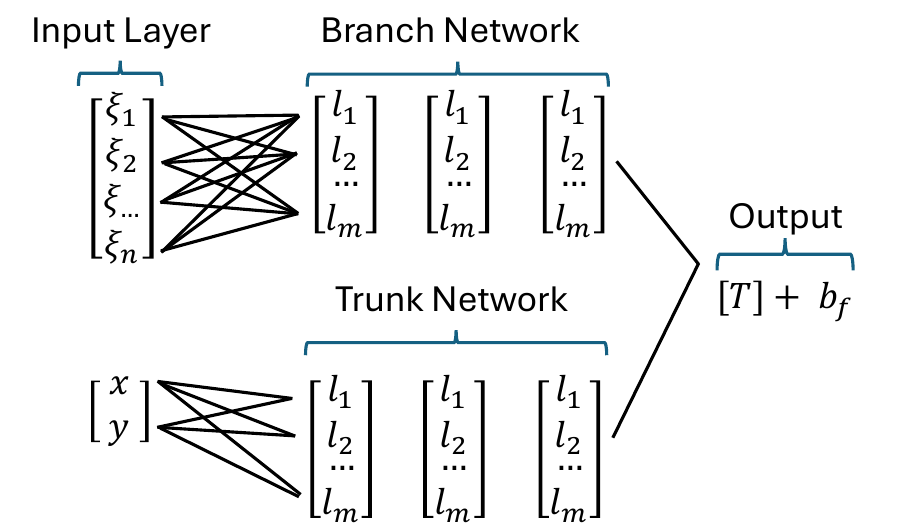} 
		\caption{PCA DeepONet Structure}
		\label{Deep_structure}
	\end{subfigure}
	\caption{\ktaddrem{Schematic of the neural network structures for (a) the fully connected neural network and (b) the PCA DeepONet. The bias vector of the last layer on both networks is denoted as $b_f$ to indicate that it is not a trainable variable but instead is fixed to the mean value of the prescribed Dirichlet boundary conditions. Values of the hyperparameters such as the number and width of each layer can be found in Table \ref{Hyperparameters}.}}
	\label{Network Structures}
\end{figure}

The most well-studied and fundamental network design for PINNs is the fully connected neural network (FNN). For a static boundary value problem, the FNN takes an input vector consisting of spatial information and returns the corresponding solution field value. This can be expressed as:

\begin{equation}
	\ktaddrem{T_{NN}(\boldsymbol\eta) = \mathbf{l}^{(N)} \cdot \sigma\left( \mathbf{l}^{(N-1)} \cdots \sigma\left( \mathbf{l}^{(1)} \boldsymbol\eta + \mathbf{b}^{(1)} \right) \cdots + \mathbf{b}^{(N-1)} \right) + \mathbf{b}^{(N)}}
\end{equation}
\noindent where, \ktaddrem{$T_{NN}(\cdot)$}  is the predicted solution, $\sigma(\cdot)$ is a nonlinear activation function, and each network layer ($i=1,...,N$) consists of a trainable weight matrix ($\mathbf{l}^{(i)}$) and bias vector($\mathbf{b}^{(i)}$). Adapting this structure to include stochastic elements is relatively straightforward. The input vector, $\boldsymbol\eta=\left\{\boldsymbol\xi,x,y\right\}$, is the concatenation of the stochastic parameters $\xi_i,i=1,\dots,N_d$ and the spatial coordinates $x$, $y$.  Figure \ref{FNN_structure} illustrates the FNN architecture used in this study. 

The second architecture used in this study is based on a Deep Operator Network (DeepONet) (\citeauthoryearkt{Lu_2021}) which is inspired by the universal approximation theorem for operators (\citeauthoryearkt{Chen_1995}). The basic idea behind the DeepONet architecture is to separate the learning of the input function representation from the spatial-temporal mapping by using separate branch and trunk networks which are combined via a dot product, given as:

\begin{equation}
		\ktaddrem{T_{NN}(\boldsymbol\xi,\mathbf{x})= \sum_{i=1}^{p} b_i(\boldsymbol\xi) \cdot t_i(\mathbf{x}) +b_f}
\end{equation}

\noindent where $b_i,t_i; i=1,...,p$ are the outputs of the branch and trunk networks, respectively, and  $p$ is the number of latent basis functions used. \ktaddrem{The last layer adds a bias term, $b_f$.} This approach therefore requires splitting the network input into a branch network component (which contains the input function representation) and a trunk network component (which contains the spatial-temporal coordinates). It remains a somewhat open ended problem for the best representation of the function argument to the branch network, as there are a number of proposed methods, such as point evaluations (\citeauthoryearkt{Lu_2021}), basis functions (\citeauthoryearkt{Kovachki_2023}), and latent spaces (\citeauthoryearkt{Kontolati_2024}). In this implementation, which is visualized in Figure \ref{Deep_structure}, the branch network receives the latent variables of the random field describing the source term as an input, while the trunk network receives the spatial components. The network is therefore tasked with the extra step of mapping these parameters to the source field, i.e. $\boldsymbol\xi \mapsto S(x,y)$, and thus this network structure is referred to as PCA-DeepONet.

\ktaddrem{In the examples below, the bias vector in the last layer is treated as non-trainable and its value is fixed to the prescribed Dirichlet boundary conditions for both network architectures. Doing so significantly speeds up the training process but not the accuracy.}

\subsection{Network Training} \label{NN_Loss}
Below describes the loss functions and training approach used in training for the various cases. The fully connected neural network is trained independently two ways: (1) against the finite element ground truth data and (2) with no data using a PINN-based loss function. The PCA-DeepONet is trained only using a PINN-based loss function without data. Additionally, the fully connected neural network is trained using a weighting strategy that biases the relative contributions of the training samples as a function of their contribution to the loss function. The details of the approaches are provided below.

\subsubsection{Data Driven Network (DDN) Loss} \label{Data_Driven_loss}
For data driven networks, the loss function is defined as the mean square error between the known values (i.e., the finite element nodal solution values) and the network's output, taking the form:

\begin{equation} 
	l_D = \frac{1}{N_n} \sum_{i=1}^{N_n} \left( T_{\text{DDN}}^{(i)} - T_{\text{FEA}}^{(i)} \right)^2
\end{equation}

\noindent where ${N_n}$ is the number of nodes in the finite element model. The total loss is then the mean of $l_{D}$ over the each the random samples, $N_s$, in the training data set:

\begin{equation} \label{Data_loss}
	\mathcal{L}_{D} = \frac{1}{N_s}\sum_{i=1}^{N_s} l_{D}^{(i)}
\end{equation}

This form of data driven loss had been shown successfully predict many types of BVPs (\citeauthoryearkt{Zhang_2022}).

\subsubsection{Physics Informed Loss} \label{P_loss}
The PINN-based loss function is defined using only the network's output and its derivatives. The approach taken in this work is to construct a loss function based on the residual of the weak form of the governing PDE (\citeauthoryearkt{XU_2021_WEAK}). Eq. \ref{weak} is modified by selecting an appropriate weight function and substituting the neural network output in place of the true solution. This error is defined as the residual loss, $l_r$:

\begin{equation} \label{Residual_loss_exact}
		l_r =   \int_\Omega  k \left[ \frac{\partial w}{\partial x}  \frac{\partial T_{NN}}{\partial x}  + \frac{\partial w}{\partial y}  \frac{\partial T_{NN}}{\partial y} \right ] - wS   \; d\Omega  
\end{equation}

Computing the exact integral in Eq. \ref{Residual_loss_exact} is not necessary as only an approximation of the residual error to guide the network training is needed. As such, reduced integration is employed to evaluate the integral on a set of collocation points: 

\begin{equation} \label{Residual_loss}
	l_r = \frac{1}{N_{cp}} \sum_{i=1}^{N_{cp}}  \left(   k \left[ \frac{\partial w^{(i)}}{\partial x}  \frac{\partial T_{NN}^{(i)}}{\partial x}  + \frac{\partial w^{(i)}}{\partial y}  \frac{\partial T_{NN}^{(i)}}{\partial y} \right ] - w^{(i)}S^{(i)} \right)^2 
\end{equation}

\noindent where $N_{cp}$ is the number of collocation points used and $(i)$ denotes the spatial  location of the $i^{th}$ collocation point. Once again the total residual loss is expressed as the sum of reach random sample's loss:

\begin{equation} \label{Total_Residual_loss}
	\mathcal{L}_{r} =  \frac{1}{N_s} \sum_{i=1}^{N_s} l_{r}^{(i)}
\end{equation}

In addition to the total residual loss, the total boundary loss, $\mathcal{L}_{b}$, is defined in order to account for the Dirichlet boundary conditions. This loss is defined as the squared difference between known boundary condition and the network's output averaged across a set of predefined boundary points, $N_{bp}$,  and averaged across all samples:

\begin{equation} \label{Boundary_loss}
 	\ktaddrem{\mathcal{L}_{b} = \frac{1}{N_s}\sum_{i=1}^{N_s}l_b^{(i)}= \frac{1}{N_s}\sum_{i=1}^{N_s}  \frac{1}{N_{bp}} \sum_{j=1}^{N_{bp}}  \left[T_{NN}^{(i,j)} - T_o  \right]^2 }
\end{equation}

The total loss for the PINN network is then defined as a weighted sum of these two losses:

\begin{equation} \label{Total_loss}
	\ktaddrem{\mathcal{L} = \mathcal{L}_{r} + w_b * \mathcal{L}_{b}}
\end{equation}
\noindent where $w_b$ is the hyperparameter used to normalize the relative magnitudes of the residual and boundary losses (see Table \ref{Hyperparameters}). This loss function is similar to those used in \citeauthoryearkt{XU_2021_WEAK}.

\subsubsection{Weighted Loss Training} \label{Weighted_loss}
The weighted loss function adopts a similar approach to cost-sensitive methods used in image classification, addressing the inherently imbalanced datasets that naturally arise when sampling from non-uniform distributions. It weighs samples more heavily to focus training on outliers in an effort to reduce the total variance of the error. This can be easily implemented through a small change in the loss function:

\begin{equation} \label{Total_loss_weighted}
	\ktaddrem{\mathcal{L}_{W} = \frac{1}{N_s}\sum_{i=1}^{N_s} W_i (l_{r}^{(i)} + w_b * l_{b}^{(i)})}
\end{equation}

\noindent where $W_i$ is a weight assigned to each sample. 

The goal is to assign higher weights to samples comprising the tails of the distribution, thereby guiding training to focus more on those samples. The approach taken is to estimate the distribution of the loss function as a function of training samples and weight the samples in contrast to their probability, such that samples with low probabilities get higher weights and vice versa for samples with higher probabilities. The effect of this is to drive the distribution of the loss function towards a uniform distribution, although in practice this is not achieved. The weighting methodology follows the approach taken by \citeauthoryearkt{Steininger_2021} given as follows. The loss function for each sample is collected in a vector $l$:
\begin{equation} 
	l_i =  \left\{ l_r^{(i)} + w_b \cdot l_b^{(i)} \right\}; i=1,...,N_s
\end{equation}

\noindent This vector can span several orders of magnitude, so it is normalized to the range \([0,1]\):

\begin{equation}
	\hat{l}_i = \frac{l_i - \min(l)}{\max(l) - \min(l)}
\end{equation}

\noindent \ktaddrem{From here the probability density function $P$ for the data $\hat{l}_i$ can be computed using a kernel density estimate (KDE)}:

\begin{equation}
	\ktaddrem{P(x)= \frac{1}{N_s h} \sum_{i=1}^{N_s} K\left( \frac{x - \hat{l}_i}{h} \right)}
\end{equation}

\noindent where $K$ is a Gaussian kernel and $h$ is the bandwith. Using this estimate for the probability density function, the weight for each sample is assigned as:

\begin{equation} \label{eq_W_2}
	\ktaddrem{W_i = \max\left( \epsilon, \; 1 - \alpha P(\hat{l}_i) \right)}
\end{equation}

\noindent where $\epsilon$ is a \ktaddrem{small threshold positive value to avoid negative values for \ktaddrem{$W_i$}}, and $\alpha$ is a scaling parameter to control the relative degree of weighting. \ktaddrem{The effect of this weight assignment strategy on the results is discussed further in Section \ref{Weighting_Resilts}}.

\subsection{Evaluation Metrics} \label{Evaluation_Metrics}
A number of error metrics are computed to evaluate the models' performances with respect to the ground truth solution, which is defined as the finite element solution. The temperature field for each of the $N_s$ realizations of the random source field is computed using 2,601 nodes arranged on a fixed grid. Thus, the ground truth finite element solutions are stored in an array of size $N_s \times 2601$. Each model's predictions are then computed at the same nodal locations and stored in a corresponding $N_s \times 2601$ model prediction array. One error metric is relative $\ell^2$-norm over the collection of all samples, given as:

\begin{equation} \label{total_L2}
	\text{Total relative } \ell^2\text{-norm} =  \frac{\| \mathbf{T}_{NN} - \mathbf{T}_{FE} \|_2}{\| \mathbf{T}_{FE} \|_2}
\end{equation}

\noindent  where $T_{NN}$ is the neural network model prediction array and $T_{FE}$ is the corresponding ground truth array. Another evaluation metric of interest is the average relative $\ell^2$-norm over all the samples, given as:

\begin{equation} \label{average_L2}
	\text{Mean relative } \ell^2\text{-norm} =  \frac{1}{N_s} \sum_{n=1}^{N_s}  \frac{\| \mathbf{T}_{NN}(n,:) - \mathbf{T}_{FE}(n,:)  \|_2}{\| \mathbf{T}_{FE}(n,:)  \|_2}
\end{equation}

\noindent Eq. \ref{average_L2} is the same evaluation metric used by \citeauthoryearkt{Wang_2021} in their study involving stochastic PDEs.

\ktaddrem{In order to more closely examine the contribution of outliers to the error metric, higher order norms than the $\ell^2$-norm are considered for model evaluation. This study, therefore, also examines the $\ell^\infty$-norm, or the maximum per-sample error (MPSE), which is defined as}:
	
	\begin{equation} \label{Peak Error_Sample}
		\textrm{MPSE}^{(i)} = \max_{j} \left| T_{\text{NN}}(i,j) - T_{\text{FE}}(i,j) \right|
	\end{equation}

\noindent and the maximum error (ME), which is defined as:
	
\begin{equation} \label{Maxium_absolute_model}
	\mathrm{ME} = \max_{i,j} \left| T_{\text{NN}}(i,j) - T_{\text{FE}}(i,j) \right|
\end{equation}

\ktaddrem{Since the $\ell^\infty$-norm is discontinuous and potentially sensitive to numerical error, continuous, higher order norms were also examined in the numerical results presented below. Specifically, the $\ell^4$, $\ell^6$, and $\ell^8$-norms corresponding to the above metrics were evaluated for all studies. The trends between uncertain input samples and model error as well as the trends among the different neural network surrogate models were consistent for all error metrics considered. For brevity, primarily the $\ell^\infty$-norm error metric is presented in the results below.}

Where appropriate, the percent error for both ME and MPSE is calculated relative to the maximum change in their thermal fields. For the ME this is defined as the difference between the maximum ground truth temperature value across all samples and the prescribed boundary condition and for MPSE this is defined as the difference between the maximum ground truth temperature of the sample and its prescribed boundary condition.

\section{Results} \label{Results}
The 2000 random samples of each source field generated in Section \ref{Source_generation} are divided into 500 samples reserved for training data and 1500 samples for test data.  The following sections present the results of various studies conducted to evaluate the relative performance of each model. The different model scenarios (i.e., combinations of networks and training procedures) are trained using identical training data to eliminate any potential bias created via random sampling. Additionally, training conducted repeatedly over multiple initial random seeds of the network parameters confirm that the sensitivity of the models' convergence and accuracy to initial values of the network parameters is insignificant for this comparative study. For brevity only the results from one representative training are presented for each model scenario.

\subsection{Hyperparameter Tuning and Benchmarking} \label{Hyper-parameters}

\begin{table}[ht]
	\centering
	\caption{Hyperparameters used for training the DDN, PINN, and PCA-DeepONet models.}
	\label{Hyperparameters}
	\renewcommand{\arraystretch}{1.2}
	\begin{tabular}{lccc}
		\toprule
		\textbf{Hyperparameter} & \textbf{DDN} & \textbf{PINN} & \textbf{PCA-DeepONet} \\
		\midrule
		Input Layer(s)             & 6x1 Single Input & 6x1 Single Input & 4x1 Branch, 2x1 Trunk \\
		& (4 stoch.+ 2 spatial)  & (4 stoch.+ 2 spatial) & (4 stoch.), (2 spatial) \\
		Hidden Layers Type         & Fully Connected & Fully Connected & Branch \& Trunk Structure  \\
		Activation Functions         &Hyperbolic Tangent & Hyperbolic Tangent & Hyperbolic Tangent \\
		Network Depth              & 4 layers        & 4 layers        & 6 (Branch), 6 (Trunk) \\
		Layer Size                 & 100             & 100             & 100 \\
		Boundary Loss Weight       & N/A             & $1 \times 10^{7}$ & $1 \times 10^{7}$ \\
		Learning Rate              & $2 \times 10^{-3}$ & $2 \times 10^{-3}$ & $2 \times 10^{-3}$ \\
		Decay Rate                 & 0.985           & 0.985           & 0.985 \\
		Decay Step                 & 500 iterations  & 500 iterations  & 500 iterations \\
		Number of Collocation Points & 2601          & 2304            & 2304 \\
		Number of Boundary Points  & N/A             & 296            & 296 \\
		\bottomrule
	\end{tabular}
\end{table}

\ktaddrem{The performance of neural networks is known to be highly dependent on hyperparameter selection, a factor that introduces significant challenges when comparing different architectures. To facilitate a fair and rigorous comparison, a consistent framework is necessary to mitigate the variability introduced by model tuning. To this end, three unifying principles are adopted to guide the comparative analysis. First, to ensure a consistent training and evaluation environment, all models are trained using identical inputs, an equivalent number of collocation points, and a common set of testing data. Second, all models are trained for the same number of epochs. Third, to validate the practical efficacy of the models, they are subjected to both quantitative and qualitative evaluations against a set of deterministic baselines. This process ensures that each model is capable of accurately generating solution fields for the specific problem under investigation.}

\ktaddrem{Rigorous testing, guided by these principles, led to the selection of hyperparameters outlined in Table \ref{Hyperparameters}. All models use the same hyperbolic tangent activation function, learning rate optimizer, and are trained on approximately 2600 individual data points per sample; the DDN network is trained on a $51 \times 51$ regular mesh, while the PINN and PCA-DeepONet models use 2304 evenly spaced interior collocation points and 296 boundary points. The batch size for each training is the full training data set. The Adam optimizer (\citeauthoryearkt{kingma2014adam}) with default parameter settings is the stochastic gradient descent (SGD) algorithm utilized to optimize the neural networks within the Tensorflow software package (\citeauthoryearkt{abadi2016tensorflow}). Each model is trained for 75,000 epochs, which is a relatively long training period to ensure training convergence. An early stopping criterion has the potential to add another source of variability when performing comparative analysis, such as sensitivity to learning rate, model size, initial parameterization, etc. For example, the loss function decreases smoothly for data driven networks compared to PCA-DeepONets in the early stages of training. Therefore, the long training duration ensures a fair comparison among the models.}

\ktaddrem{One significant difference among model parameters is necessary. PINNs are typically trained with 3 to 4 fully connected hidden layers (\citeauthoryearkt{RAISSI_2019}, \citeauthoryearkt{Zhang_2020}), while DeepONets have a branch and trunk structure that typically use only 2 to 3 hidden layers each (\citeauthoryearkt{Lu_2021}, \citeauthoryearkt{Kontolati_2024}), and both use different layer widths which appear to be problem dependent. Changing the number of hidden layers or the layer size has a significant impact on the model size (i.e., the total number of trainable parameters) and affects training time; however, forcing a standardization overly penalizes a given model, so an allowance is necessary. During the qualitative evaluations of the deterministic baselines, the PCA-DeepONet performs noticeably better with a network depth of 6 layers compared to 4, while the DNN and PINN networks do not show any appreciable improvement in accuracy beyond 4 layers. Hence, 6 layers are used for the PCA-DeepONet and 4 layers are used for the DNN and PINN networks.}

\begin{figure}[ht]
	\centering
	\includegraphics[width=1\textwidth]{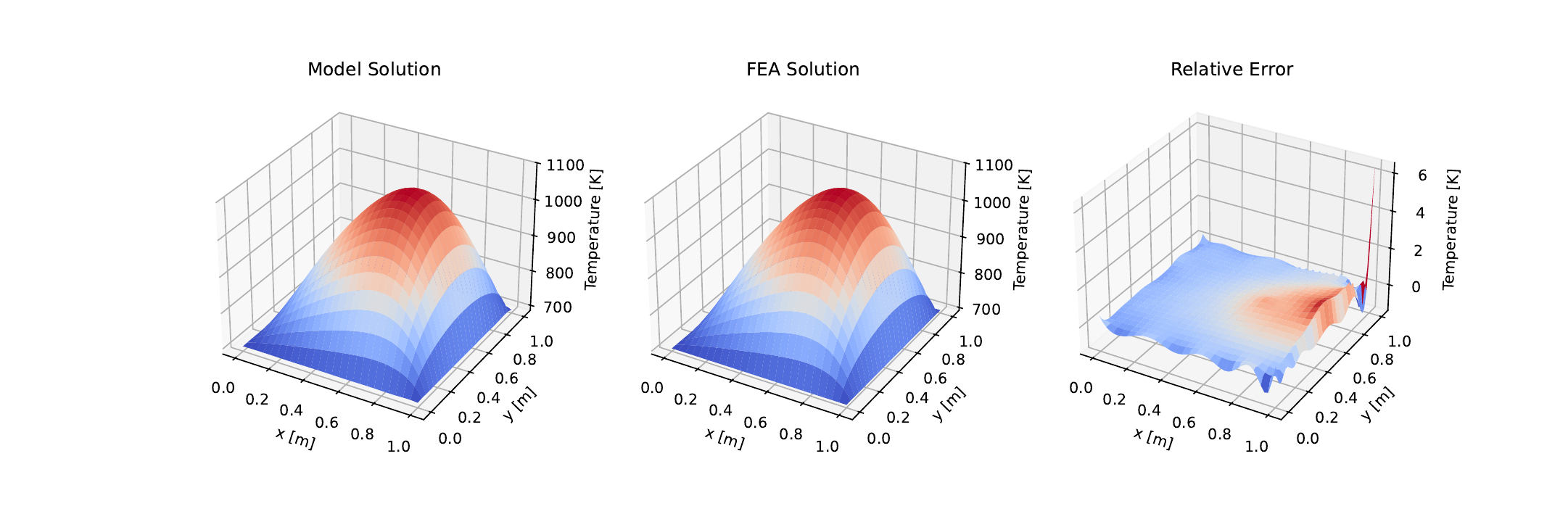}
	\caption{The PINN model benchmark for sample $\#196$ from Source 2 has the highest ME of all benchmark tests. The error is largely localized to the boundary with a sharp temperature peak occurring at the corner boundary of the domain.}
	\label{PINN_benchmark}
\end{figure}

\ktaddrem{Table \ref{Benchmark_Data_With_Tmax} and Figure \ref{PINN_benchmark} present the results of the final benchmark tests. These results demonstrate that the selected hyperparameters and network structures} can solve individual PDEs to a sufficient level of accuracy. It is important to note that the hyperparameter values in Table \ref{Hyperparameters} are tuned to minimize ME (as defined in Eq. \ref{Maxium_absolute_model}) on training datasets containing 200 samples, not just the benchmark tests. This explains why there is measureable error in some of the benchmarks. The worst performing benchmark is the PINN model trained on sample $\#196$ from Source 2 (which featured the maximum temperature among all samples from Source 2) and whose MPSE is 6.37 or $1.65\%$.

\ktaddrem{Lastly, the models are trained on AMD 9654 Genoa and AMD 7H12 Rome CPUs. Because of differences in model size, code optimization, and training hardware, direct comparison of the computation efficiency of each model are not made. However, the DDN is generally the most efficient, followed by the PINN, and then the PCA-DeepONet.}

\begin{table}[ht]
	\centering
	\caption{The maximum temperature, total relative $\ell^2$ error, and maximum per-sample error (MPSE), for different models across four benchmark samples.}
	\label{Benchmark_Data_With_Tmax}
	\begin{tabular}{|l|c|cc|cc|cc|}
		\hline
		&  & \multicolumn{2}{c|}{\textbf{DDN}} & \multicolumn{2}{c|}{\textbf{PINN}} & \multicolumn{2}{c|}{\textbf{PCA-DeepONet}} \\
		\textbf{Source - Sample \#} & $T_{\text{max}}$ (K) & $\ell^2$ & MPSE & $\ell^2$ & MPSE & $\ell^2$ & MPSE \\
		\hline
		Source 1 - \#17  & 783.4 & 3.98e-5 & 0.11 & 3.98e-5 & 0.11 & 7.62e-5 & 0.35 \\
		Source 1 - \#26  & 730.3 & 3.72e-5 & 0.15 & 1.61e-5 & 0.13 & 7.83e-5 & 0.16 \\
		Source 2 - \#196 & 1108.8 & 4.81e-4 & 3.48 & 7.70e-4 & 6.37  & 4.54e-4 & 1.12 \\
		Source 2 - \#43  & 852.7 & 1.27e-4 & 1.04 & 1.79e-4 & 0.49 & 5.24e-4 & 1.60\\
		\hline
	\end{tabular}
\end{table}

\subsection{Training Data Size Analysis}
For stochastic neural networks, incorporating a larger number of training samples should enhance predictive accuracy and provide a more comprehensive understanding of the system. This improvement comes at a cost, as training models on more samples increases computational demands. Training a deterministic PINN can take anywhere from a few minutes to several hours, depending on the type of PDE, number of spatial dimensions, domain size, geometry, complexity of the loss function, and whether GPU acceleration is used. When combined with the need to evaluate hundreds (or more) of training samples, the total training time can become a significant consideration. Since the focus of this study is on model accuracy rather than computational efficiency, all models are compared after a fixed number of training epochs rather than a fixed training time. This allows performance trends to be evaluated without penalizing the additional time required to train on larger datasets.

\begin{figure}[ht]
	\centering
	\includegraphics[width=1\textwidth]{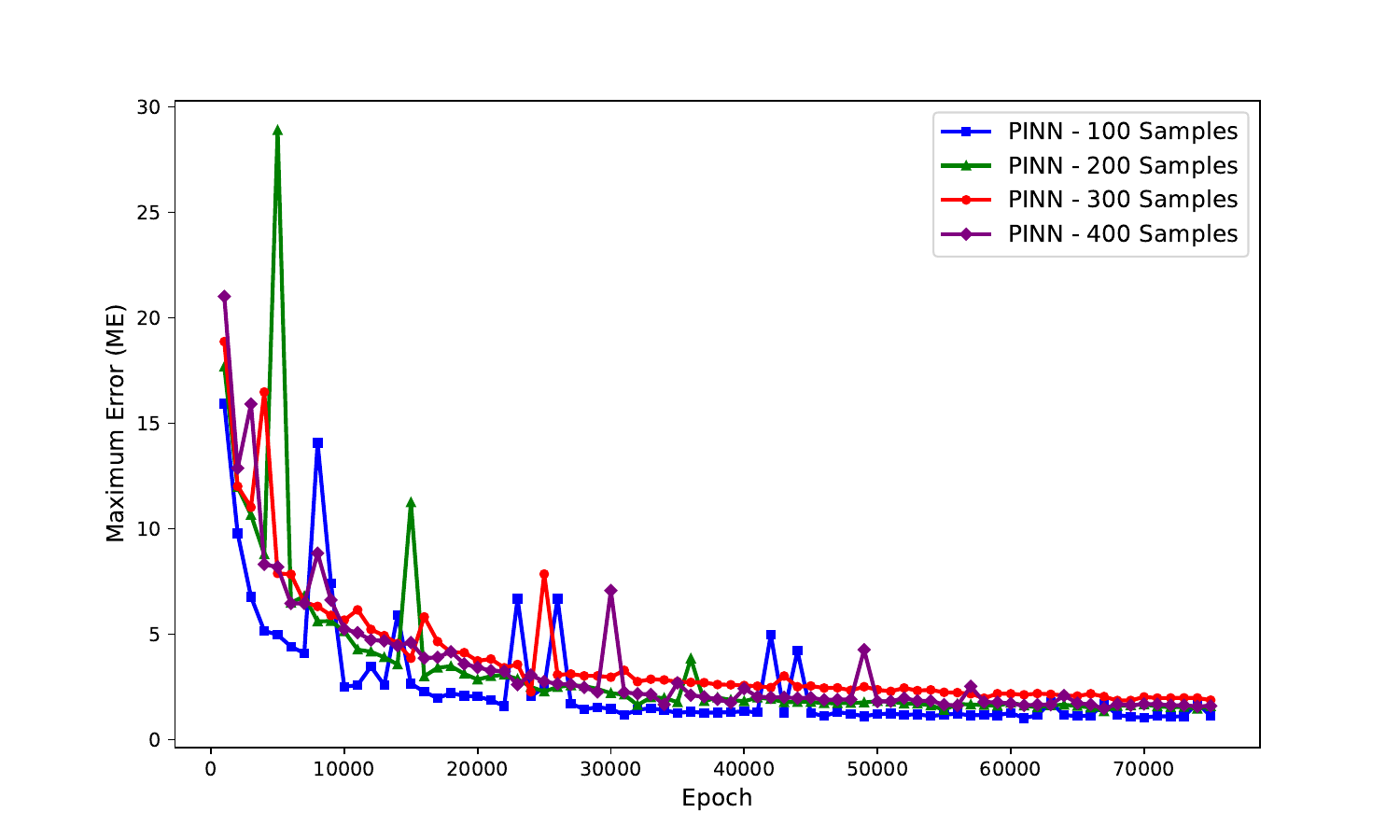}
	\caption{Convergence history of the maximum error (ME) during training for PINNs trained on an increasing number of random samples.  Increasing the training data from 100 to 400 samples increases the training time by a factor of about 2.}
	\label{Num_samples_convergence}
\end{figure}

Figure \ref{Num_samples_convergence} shows the convergence behavior of PINN models trained with an increasing number of random samples drawn from Source 1. The overall convergence rates are quite similar across all models, with each model reaching a training maximum error (ME) between $2$ and $3~K$ after 75,000 epochs. Table \ref{Num_samples_table} summarizes the final training and testing errors for each model. For all model types, increasing the number of training samples had only a nominal impact on the training errors. Slight increases in ME are observed, which in most cases could be attributed to the addition of a new sample with a maximum temperature exceeding that of all previous samples. 

\begin{table}[h]
	\centering
	\caption{Training and testing, Total relative $\ell^2$ error and maximum error (ME) for DDN, PINN, and PCA-DeepONet models trained on varying sample sizes across Source 1 and Source 2. Errors for each source are computed on a fixed set of 1500 testing samples.}
	\label{Num_samples_table}
	\resizebox{\textwidth}{!}{%
		\begin{tabular}{lcccccccc}
			\hline
			\textbf{Model} & \multicolumn{4}{c}{\textbf{Source 1}} & \multicolumn{4}{c}{\textbf{Source 2}} \\
			($\#$ of Samples)& Train $\ell^2$ & Train ME & Test $\ell^2$ & Test ME & Train $\ell^2$ & Train ME & Test $\ell^2$ & Test ME \\
			\hline
			DDN (100)  &$3.99\mathrm{e}{-5}$  & 1.4 & $1.18\mathrm{e}{-2}$ & 71.45 & $2.19\mathrm{e}{-4}$ &5.37 &$4.73\mathrm{e}{-2}$ &334.21  \\
			DDN (200)  &$4.89\mathrm{e}{-5}$  & 1.52 & $2.15\mathrm{e}{-3}$ &30.36 &   $1.67\mathrm{e}{-4}$ &7.68 &$5.92\mathrm{e}{-3}$ &117.86  \\
			DDN (300)  &$5.54\mathrm{e}{-5}$  & 1.91 & $4.47\mathrm{e}{-4}$ & 15.06 &   $2.02\mathrm{e}{-4}$ & 8.43 &$2.83\mathrm{e}{-3}$ & 72.10 \\
			DDN (400)  & $4.40\mathrm{e}{-5}$ & 1.63 &  $4.40\mathrm{e}{-4}$&  15.93 & $4.11\mathrm{e}{-5}$ & 9.92 &$1.69\mathrm{e}{-3}$  & 60.04 \\
			\hline
			PINN (100) & $7.05\mathrm{e}{-5}$ & 2.13 &  $5.45\mathrm{e}{-3}$&  61.09 & $7.21\mathrm{e}{-5}$ & 10.91 &$2.98\mathrm{e}{-2}$  & 292.63\\
			PINN (200) & $5.56\mathrm{e}{-5}$ & 2.75 &  $6.25\mathrm{e}{-4}$&  13.19 & $5.02\mathrm{e}{-4}$ & 16.11 &$3.89\mathrm{e}{-3}$  & 76.63\\
			PINN (300) &$4.35\mathrm{e}{-5}$  & 3.17 &$2.65\mathrm{e}{-4}$   & 10.08 & 
			$4.77\mathrm{e}{-4}$ & 20.44 &$1.43\mathrm{e}{-3}$  & 38.60\\
			PINN (400) & $6.36\mathrm{e}{-5}$ & 2.85 &  $2.71\mathrm{e}{-4}$& 10.48 & 
			$4.57\mathrm{e}{-4}$ & 17.94 &$9.34\mathrm{e}{-4}$  & 29.48 \\
			\hline
			PCA-DeepONet (200) & $6.00\mathrm{e}{-5}$  & 1.13 &$6.67\mathrm{e}{-3}$   & 44.26 & 
			$2.04\mathrm{e}{-4}$ & 3.66 &$2.87\mathrm{e}{-2}$  & 273.84\\
			PCA-DeepONet (400) & $1.12\mathrm{e}{-4}$ & 0.99 &$2.50\mathrm{e}{-3}$  & 26.85& 
			$7.64\mathrm{e}{-4}$ & 16.63 &$1.66\mathrm{e}{-2}$  & 210.88\\
			\hline
		\end{tabular}%
	}
\end{table}

\begin{figure}[ht]
	\centering
	\includegraphics[width=1\textwidth]{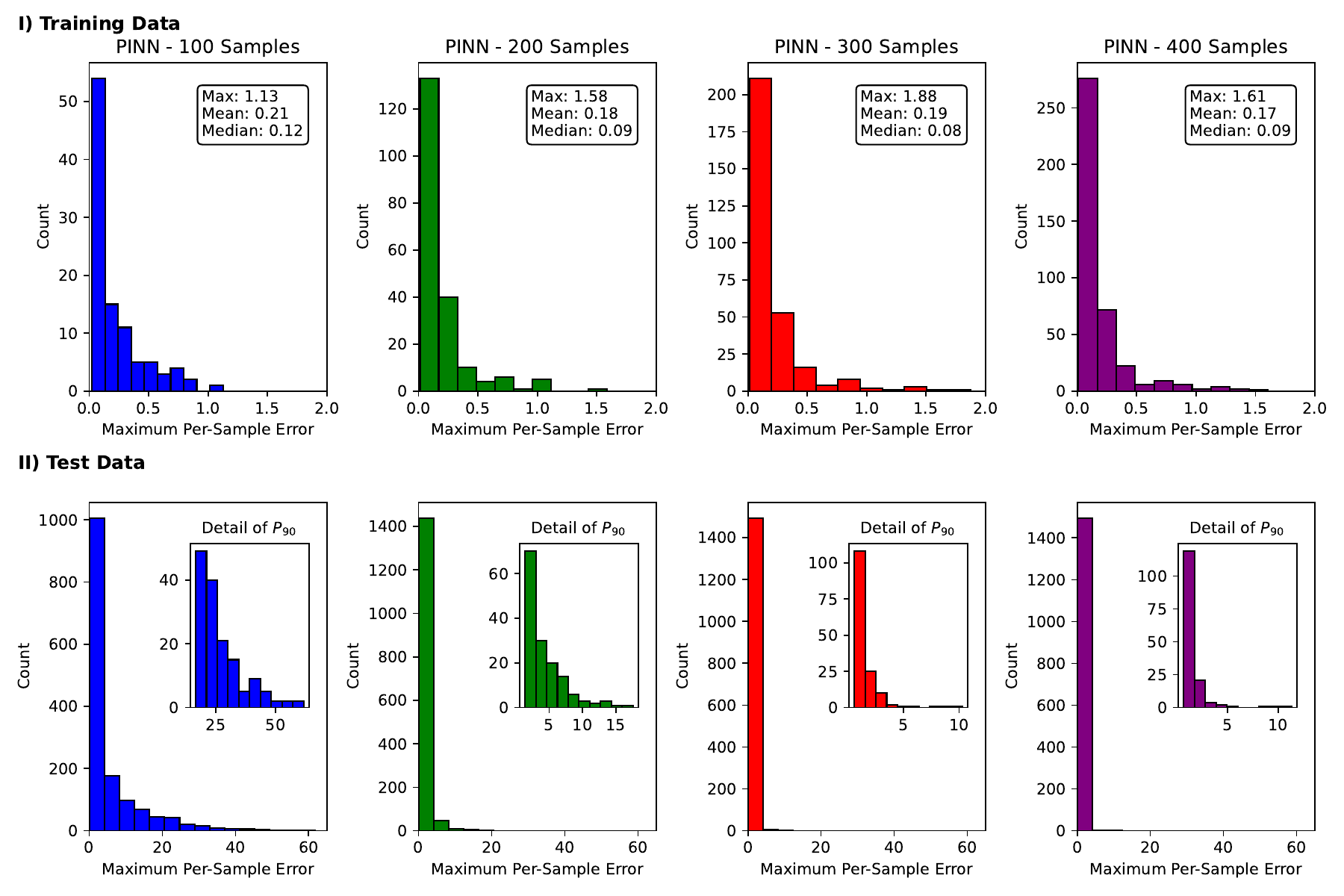}
	\caption{Histograms of the MPSE of the PINN model for (a) training data, and (b) test data for increasing number of samples. Increasing the number of samples does not have a significant effect on MPSE of the training data (subfigure (a)), while it dramatically decreases the MPSE of the testing data (subfigure (b)). For example, an 83.3\% drop in the MPSE is achieved by increasing the training data from 100 to 300 samples. However this effect plateaus after 300 samples as both the 300 sample and 400 sample data sets have the approximately the same maximum model error.}
	\label{Num_samples_hist}
\end{figure}

On the other hand, the generalization error improves significantly as more training samples are added. Both the DDN and PINN models exhibit very high ME on the testing data when trained with only 100 samples. Increasing the number of training samples from 100 to 200 results in at least a $50\%$ reduction in ME across all trials. However, this effect is not limited to just the worst-performing samples. Figure \ref{Num_samples_hist} shows histograms of the training and testing MPSE, where it can be seen that there is significant improvement across nearly all samples.

Figure \ref{Num_samples_hist} and Table \ref{Num_samples_table} also show that the magnitude of this improvement diminished with additional samples. This is expected because it is increasingly less likely that each new sample provides substantially new information since random sampling is used to generate the training data. In the case of Source 1, this is reflected in the maximum temperature across all samples in the training data: it increases from 100 to 200 samples and from 200 to 300, but not from 300 to 400. 

\begin{figure}[ht]
	\centering
	\includegraphics[width=1\textwidth]{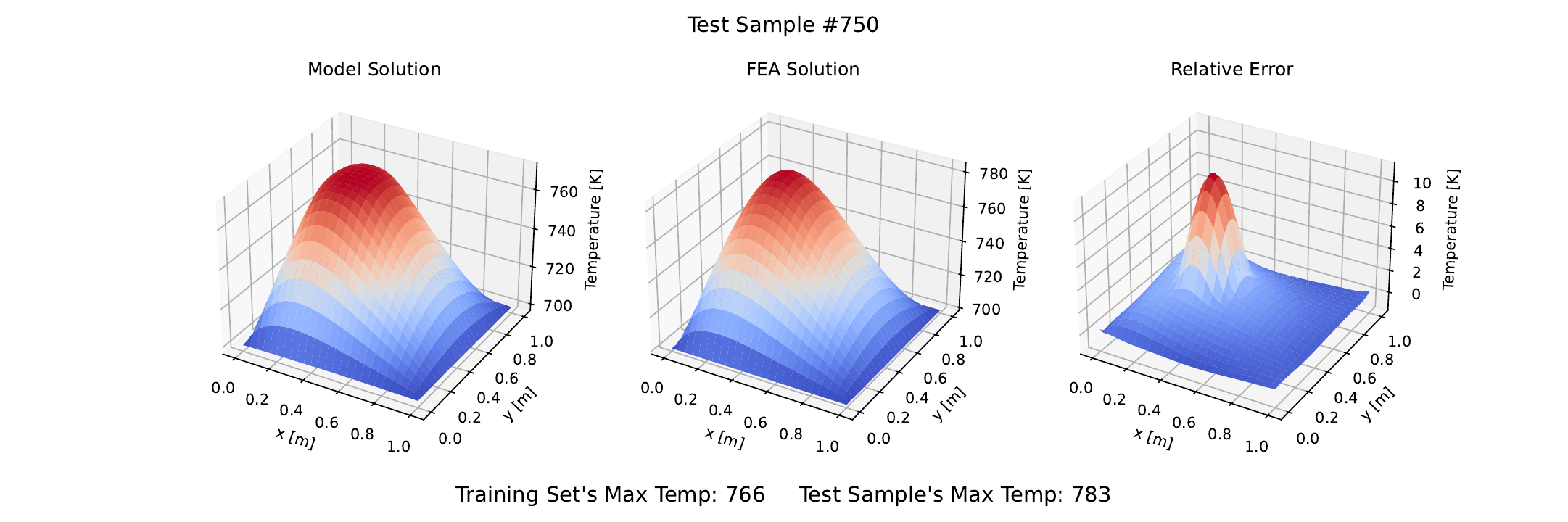}
	\caption{The predicted solution field from Source 1 with the greatest MPSE, where the true temperature is underestimated by approximately $10~K$. This sample is particularly challenging to capture, as it requires the model to extrapolate beyond the training dataset because its peak temperature is $17~K$ larger than the maximum temperature value in the training data set.}
	\label{Npt_750}
\end{figure}

Ultimately, the maximum temperature observed in the training data plays a critical role, as it effectively sets a soft upper limit on model predictions. This relationship is illustrated in Figure \ref{Npt_750}. The ground truth for the highlighted sample has a maximum temperature of 783 $K$, approximately 17 $K$ higher than any value encountered during training. A visual inspection of the model's predicted solution shows a flattened temperature peak in this region. Correspondingly, the relative error plot shows that the model's maximum error occurs in this region, underscoring the models difficulty in accurately extrapolating beyond its training range. The model is able to partially extrapolate the temperature only 5.7 $K$ vs the 17 $K$ required.  However, this sample and other extrapolated samples (see figure \ref{Scoure_1_classified}) are the largest sources of error and lead to the general plateauing of the generalization error between 300 and 400 training samples seen in Table \ref{Num_samples_table}. 

Extrapolation is a known challenge for neural networks in general, and it should not be surprising that these neural networks struggle in this area. The stochastic network needs to be trained on a sufficient number of samples such that the probability of a worse sample becomes sufficiently small. For a relatively small number of stochastic dimensions (such as the four considered in this paper), a random sampling approach is feasible. However, for a larger number of stochastic dimensions, it's likely that a smarter sampling strategy would need to be implemented to create a more optimized training dataset. The success of such a method would partly depend on the nature of the random field and whether there is enough correlation between variables to guide where additional samples should be placed. \ktaddrem{This is further discussed in Section \ref{Individual Sample Analysis}.}

\subsection{Comparing Model Types} 
\begin{figure}[ht]
	\centering
	\includegraphics[width=1\textwidth]{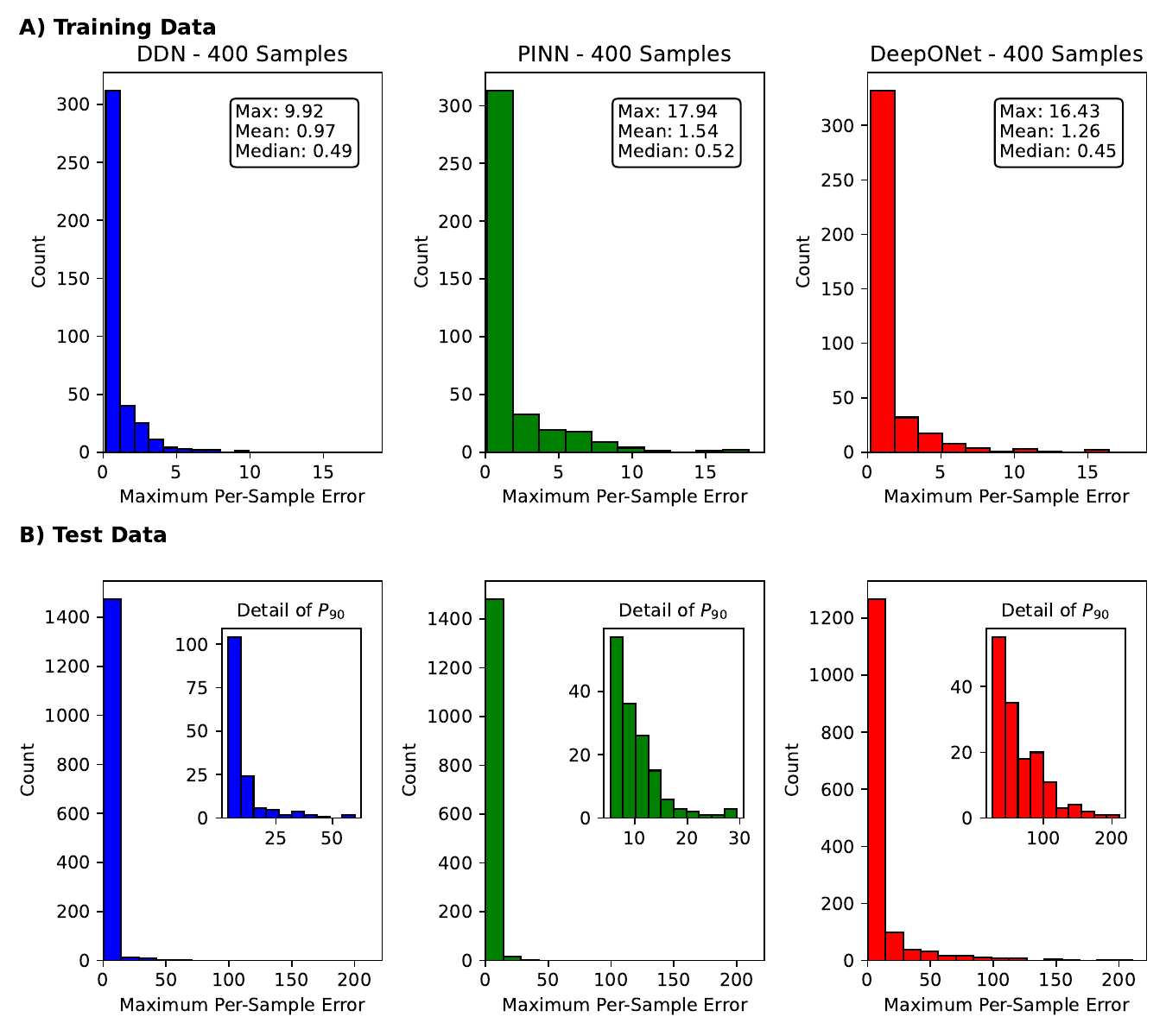}
	\caption{Maximum per-sample errors (MPSE) histograms for each models' (a) training data and (b) testing data.}
	\label{Comparison_hist}
\end{figure}

In this subsection, the performance of each of the model types is compared and contrasted to better understand the performance trade offs of each. The data shows that the DDN network outperforms the other two models on the training data and the PINN network performs best on the testing data in terms of reducing ME and the distribution of MPSE values. Figure \ref{Comparison_hist} shows each models' performance on a 400 sample training set from Source 2. The ME for PINN network is 29.48 vs. 60.04 for the DDN, and 210.88 for the PCA-DeepONet. Table \ref{tab:combined_stats_narrow} contains the counts of the total number of MPSEs that are above certain percentage thresholds and the average MPSE of each model. It shows that despite having a slightly lower average MPSEs, the DDN error has more individual samples whose MPSE exceeded $>4\%$, meaning the distribution of MPSE values has a much heavier upper tail for the DDN compared to the PINN.

\begin{table}[ht]
	\centering
	\caption{Sample error statistics for models trained on 400 samples. The table shows the count of samples whose MPSE exceeded specific percentage error thresholds (relative to the maximum temperature change of 83.4 K for Source 1 and 408.6 K for Source 2) and the average MPSE across training and testing sets.}
	\label{tab:combined_stats_narrow}
	\begin{tabular}{ll cccc cc}
		\hline
& & \multicolumn{4}{c}{\textbf{Samples with MPSE $> $ Threshold}} & \multicolumn{2}{c}{\textbf{Mean MPSE (\%)}} \\
		\cline{3-6} \cline{7-8}
		\textbf{Source} & \textbf{Model} & \textbf{Train $>2\%$} & \textbf{Test $>2\%$} & \textbf{Test $>4\%$} & \textbf{Test $>6\%$} & \textbf{Train} & \textbf{Test} \\
		\hline
		1 & DDN & 0 & 61 & 27 & 18 & 0.22 & 0.42 \\
		1 & PINN & 6 & 85 & 22 & 12 & 0.26 & 0.41 \\
		1 & PCA-DeepONet & 0 & 305 & 186 & 143 & 0.18 & 1.48 \\
		\hline
		2 & DDN & 1 & 54 & 20 & 14 & 0.97 & 1.70 \\
		2 & PINN & 12 & 84 & 12 & 4 & 1.54 & 1.89 \\
		2 & PCA-DeepONet & 7 & 318 & 209 & 153 & 1.26 & 9.08 \\
		\hline
	\end{tabular}
\end{table}

\begin{figure}[ht]
	\centering
	\begin{subfigure}[b]{0.48\textwidth}
		\centering
		\includegraphics[width=\textwidth]{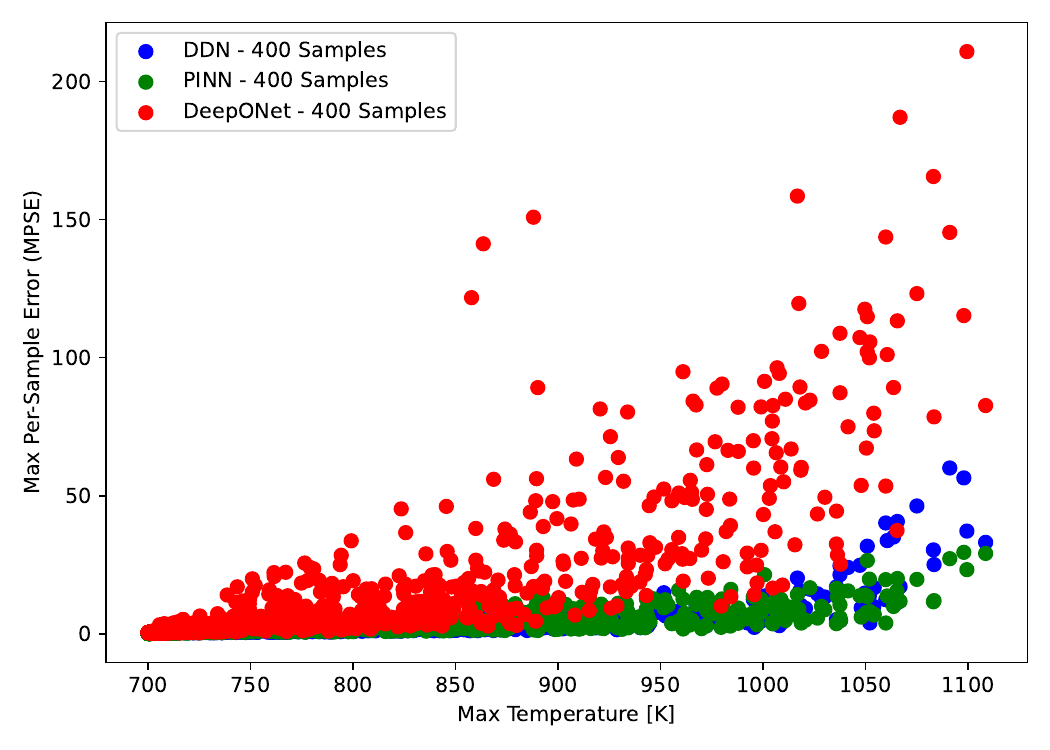}
		\caption{Maximum per-sample error (MPSE).}
		\label{ME_MPSE}
	\end{subfigure}
	\hfill
	\begin{subfigure}[b]{0.48\textwidth}
		\centering
		\includegraphics[width=\textwidth]{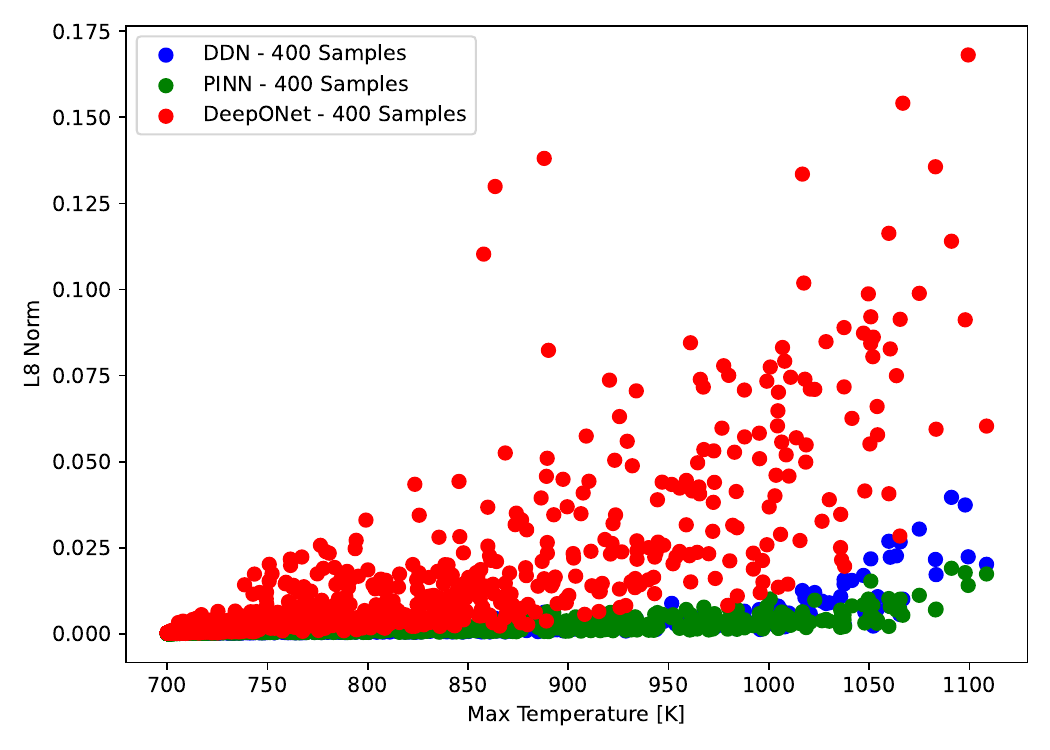}
		\caption{Relative $\ell^8$-norm}
		\label{ME_L8}
	\end{subfigure}
	\caption{\ktaddrem{The maximum ground truth temperature of each sample is plotted against its (a) maximum per-sample error (MPSE) and (b) the $\ell^8$-norm of the difference between the model predictions and the ground truth divided by the $\ell^8$-norm of the ground truth for all 1500 testing samples for Source 2. The $\ell^8$-norm is also shown to demonstrate that the trend shown in (a) is not the result of numerical sensitivity associated with maximum values. The PCA-DeepONet clearly under performs the DDN and PINN models for both error metrics.}}
	\label{Sample_Me_Scatter}
\end{figure}

This bias is further illustrated in Figure \ref{Sample_Me_Scatter} which visualizes the data according to each sample's maximum temperature. The PCA-DeepONet clearly underperforms the other two models, and while both the DDN and PINN models perform similarly on samples less the 1000 $K$, the DDN exhibits a stronger trend towards larger per-sample error for samples with larger maximum temperatures compared to the PINN.

One surprising result is the poor performance of the PCA-DeepONet on this testing data. The PCA-DeepONet performs well on the training data, even beating the PINN on both Source 1 and Source 2, and the DDN on Source 1. However, its test predictions are significantly worse than the DDN or PINN. In Figure \ref{Sample_Me_Scatter} the PCA-DeepONet performs the worst on nearly every sample. Nevertheless, it achieves a mean relative $\ell^2$ error on the testing data of just $0.56\%$ which is within the margin of error of $0.45 \pm 0.16\%$ achieved by \citeauthoryearkt{Wang_2021} on a stochastic diffusion-reaction problem, indicating that the the model is not just pooly trained as it is on par with a similar model presented in the literature. 

These results are not isolated to just a single training. Similar results hold for trainings on the data sampled from Source 1 and for training sets which contain a different number of  samples. These results provides evidence that as a surrogate model for mapping stochastic parameters to solution field values, the PINN model outperforms traditional data driven training and PCA-DeepONet on identical testing data.

\subsection{Weighted vs Unweighted Training} \label{Weighting_Resilts}
In this subsection, the cost-sensitive weighting approach described in Section \ref{Weighted_loss} is used to attempt to correct for an apparent error bias towards high temperature samples. Figures \ref{Sample_Me_Scatter} show that samples with more extreme maximum temperatures tend to have higher MPSE values. This indicates that the network training remains susceptible to the long-tail problem, and that the relatively few samples with large maximum temperatures in the training data do not provide enough information to properly train the network to handle these extreme values in test cases. 

\begin{figure}
	\centering
	\begin{subfigure}[b]{0.48\textwidth}
		\centering
		\includegraphics[width=\textwidth]{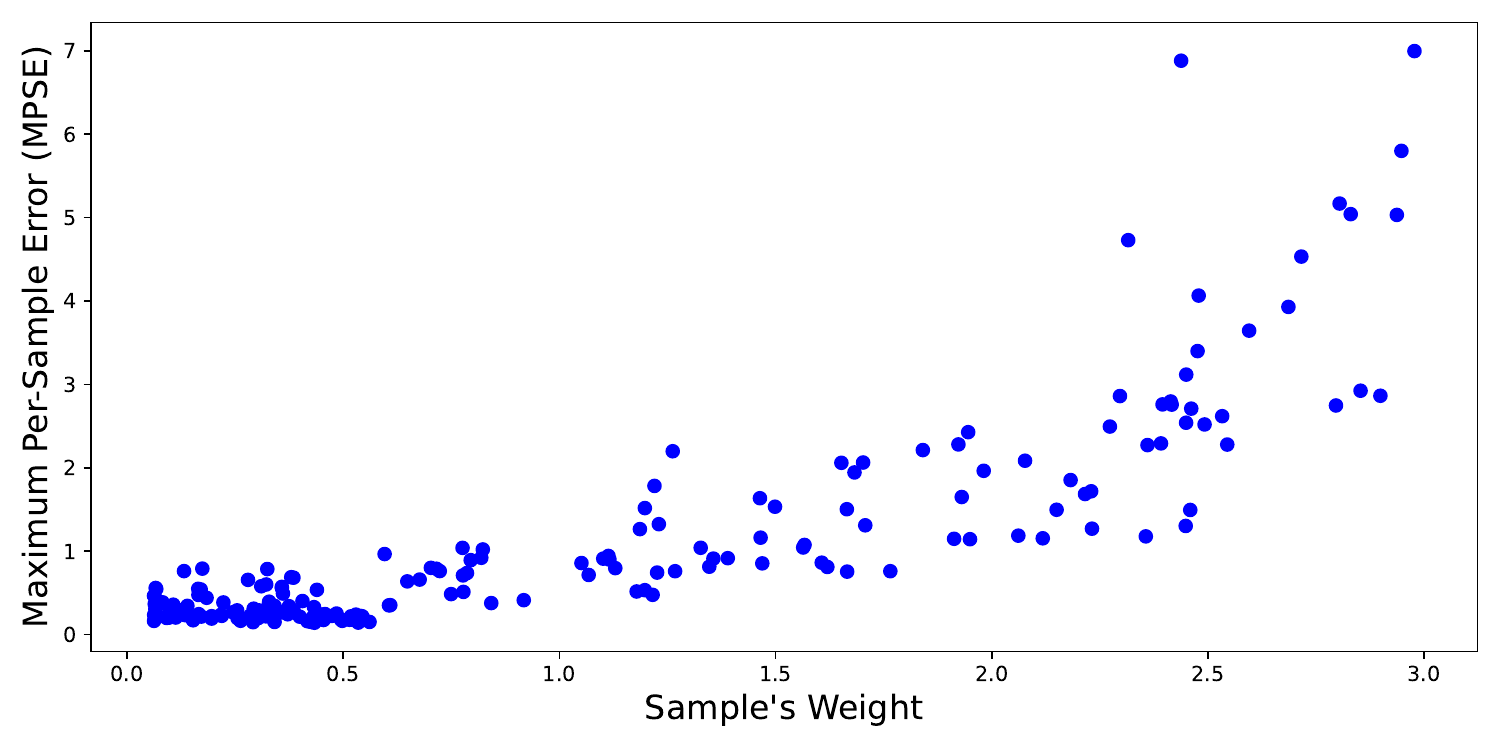}
		\caption{Sample weights vs. maximum per-sample error (MPSE).}
		\label{fig:weights_error}
	\end{subfigure}
	\hfill
	\begin{subfigure}[b]{0.48\textwidth}
		\centering
		\includegraphics[width=\textwidth]{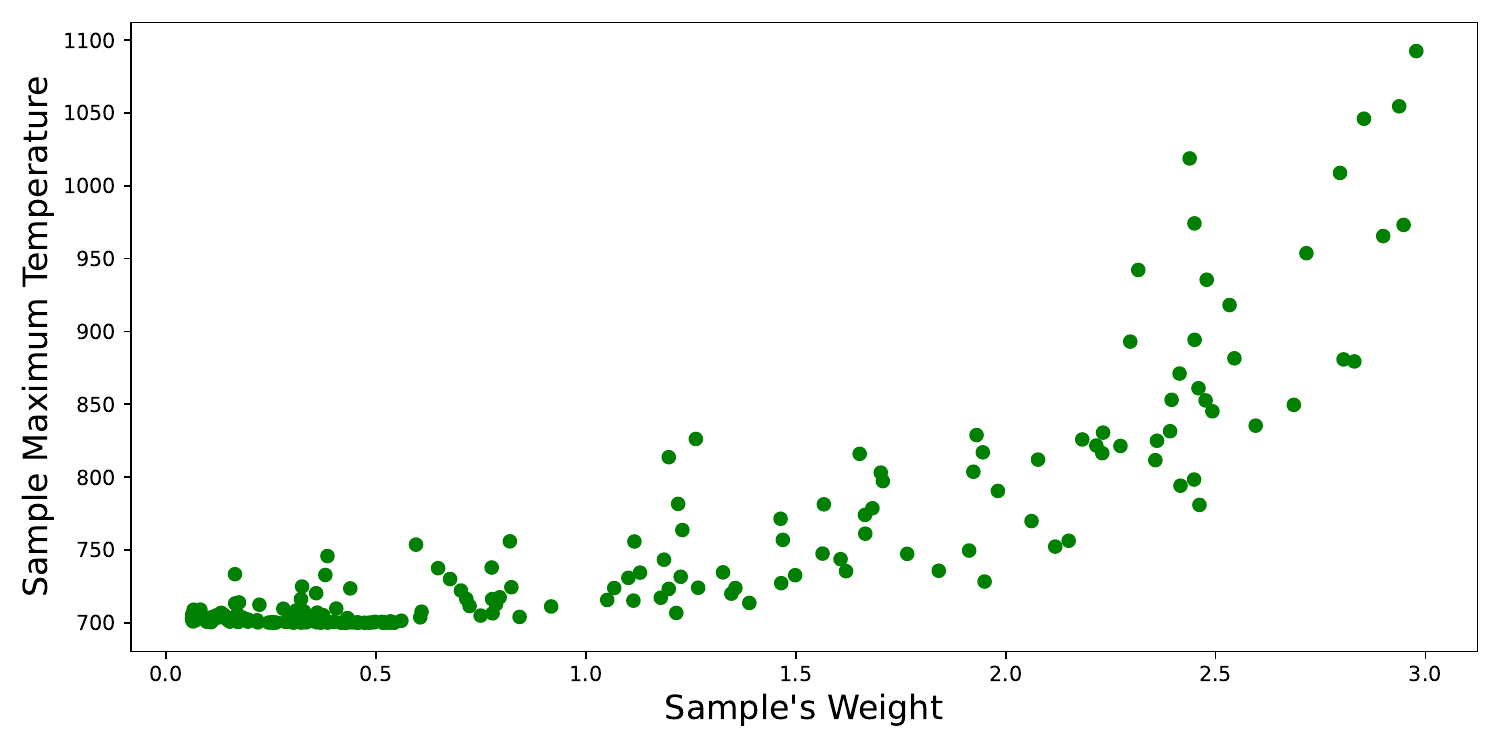}
		\caption{Sample weights vs. maximum temperatures.}
		\label{fig:weights_temperature}
	\end{subfigure}
	\caption{Relationships between sample weight and the (a) training error and (b) maximum solution field temperature for Source 2 training data. The scatter plots show the assigned weights computed after 50K epochs.}
	\label{Weights_detailed}
\end{figure}

\begin{figure}[ht]
	\centering
	\includegraphics[width=1\textwidth]{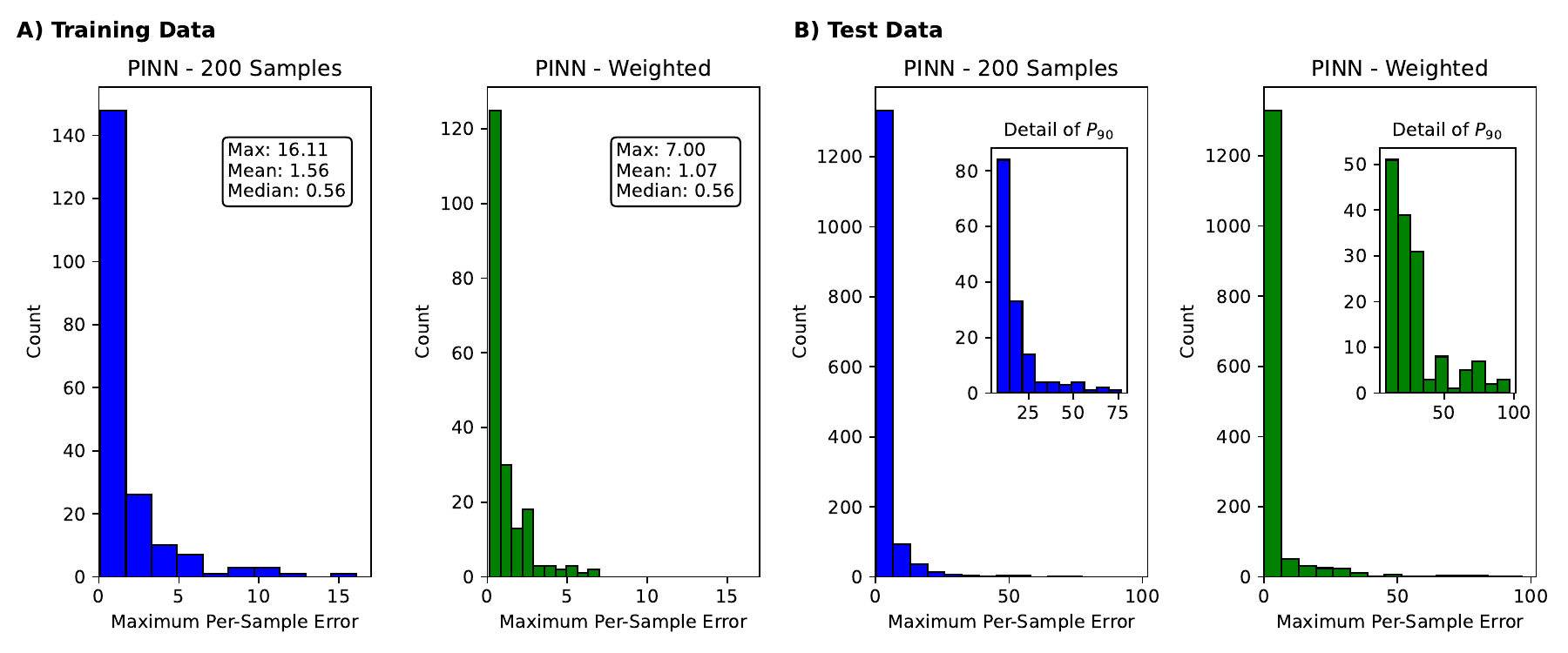}
	\caption{A comparison of the weighted vs. non-weighted PINN. Each model is trained on the same 200 training samples and each network is initialized with the same random seed. Weighting does improve performance on the training data but had an adverse effect on testing data.}
	\label{Weighted_comparison}
\end{figure}

The PINN model is retrained using a weighted loss function with a scaling value of $\alpha = 0.4$ and a minimum weight of $\epsilon = 0.05$ as defined in Eq. \eqref{eq_W_2}. As shown in Figure \ref{Weights_detailed}, the weighted loss function has the desired effect of assigning higher weights as the loss increases. There is a strong correlation between a sample's assigned weight and both its maximum temperature and its maximum sample error. This guides the training process and leads to a reduction in the training data's ME from 16.11 $K$ to \ktaddrem{7.00} $K$. However, the testing data predictions for the weighted samples worsen, with the ME increasing from 76.63 $K$ to 96.72 $K$ (See Figure \ref{Weighted_comparison}). This effect is observed even if the number of training samples is increased. The ME of the PINN model trained on a dataset of 400 samples decreases from 17.94 $K$ to 12.65 $K$ on the training data while it increases from 29.72 $K$ to 56.73 $K$ on the testing data.

There are various potential reasons why the weighted loss function does not result in significant improvements in the ME for the testing data. First, the weighting does not directly address the largest source of error, which is extrapolation in the test data. In fact, as evidenced by the improved training error, it is possible that the weighting strategy increases the model's overfitting, since the weighting improves optimization error while increasing generalization error. Second, it is unclear if there is an optimal weighting model to adaptively rebalance training to uniformly weight the entire distribution of the quantity of interest (i.e. the distribution of maximum sample temperatures). Several other weighting methods have been tested including (1) a fixed weighting strategy based on the initial residual loss of each sample, (2) a strategy that separately weights the boundary and residual loss terms based on their respective values, and (3) a ramped weighting strategy where the total weighting is increased gradually. However, results for each these methods do not improve (or are worse) compared to the presented weighting strategy, and as such their individual results are omitted for brevity. Nevertheless, given the improvement on the training data it remains possible that different weighting strategies could yield positive results and warrants further study, although it is likely the case that the best weighting strategy is data-specific and therefor not generalizable. 

\begin{figure}[ht]
	\centering
	\includegraphics[width=1\textwidth]{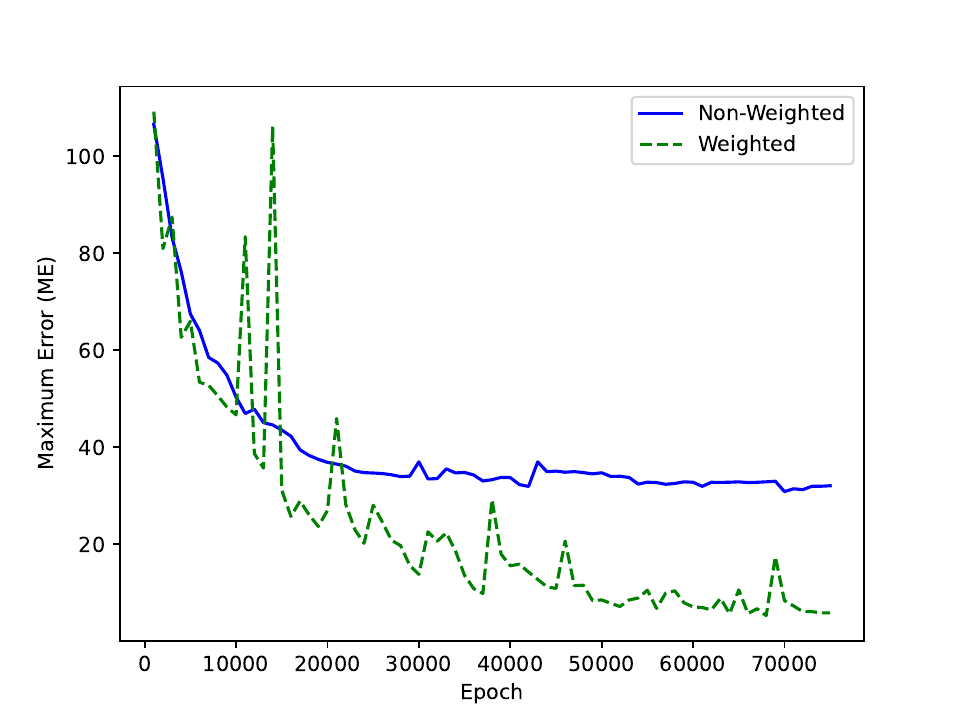}
	\caption{Convergence rates for PINN models trained on a dataset of 200 samples using both weighted and non-weighted loss functions.}
	\label{Weighted_cp8}
\end{figure}

\ktaddrem{One promising application where the weighted training provides a potential improvement to training occurs when the residual loss is estimated using a sparse grid of collocation points. To demonstrate this, the PINN model is retrained by reducing the number of collocation points per sample from 2600 to 316 (with 256 interior collocation points and 60 boundary points). In this study, the spatial variability of the source field is also increased with the goal of making the source field distribution harder to resolve with the sparse grid of collocation points. This is done by modifying its underlying correlation function defined in Eq. \ref{Auto Correlation} to $R(\tau_x,\tau_y)=\mathrm{exp}\left(-0.03(\tau_x+ \tau_y)^2\right)\cos(\tau_x) + \mathrm{exp}\left(-0.06 (\tau_x + \tau_y)^3\right)\cos(\tau_y)$. Figure \ref{Weighted_cp8} presents the results of training with and without weighting, showing extremely large discrepancy in error between the two cases. The weighted case converges toward a training error of less than 10 and the unweighted case converges to approximately 40. The testing ME also improved from 89.99 for the unweighted case down to 63.96 for the weighted case. These results suggest that weighting collocation points with reduced integration could be a computationally efficient training approach; however, further study is needed to assess the impact on compute time and accuracy.}

\subsection{Individual Sample Analysis} \label{Individual Sample Analysis}
\begin{figure}[ht]
	\centering
	\begin{subfigure}[t]{0.48\textwidth}
		\centering
		\includegraphics[width=\textwidth]{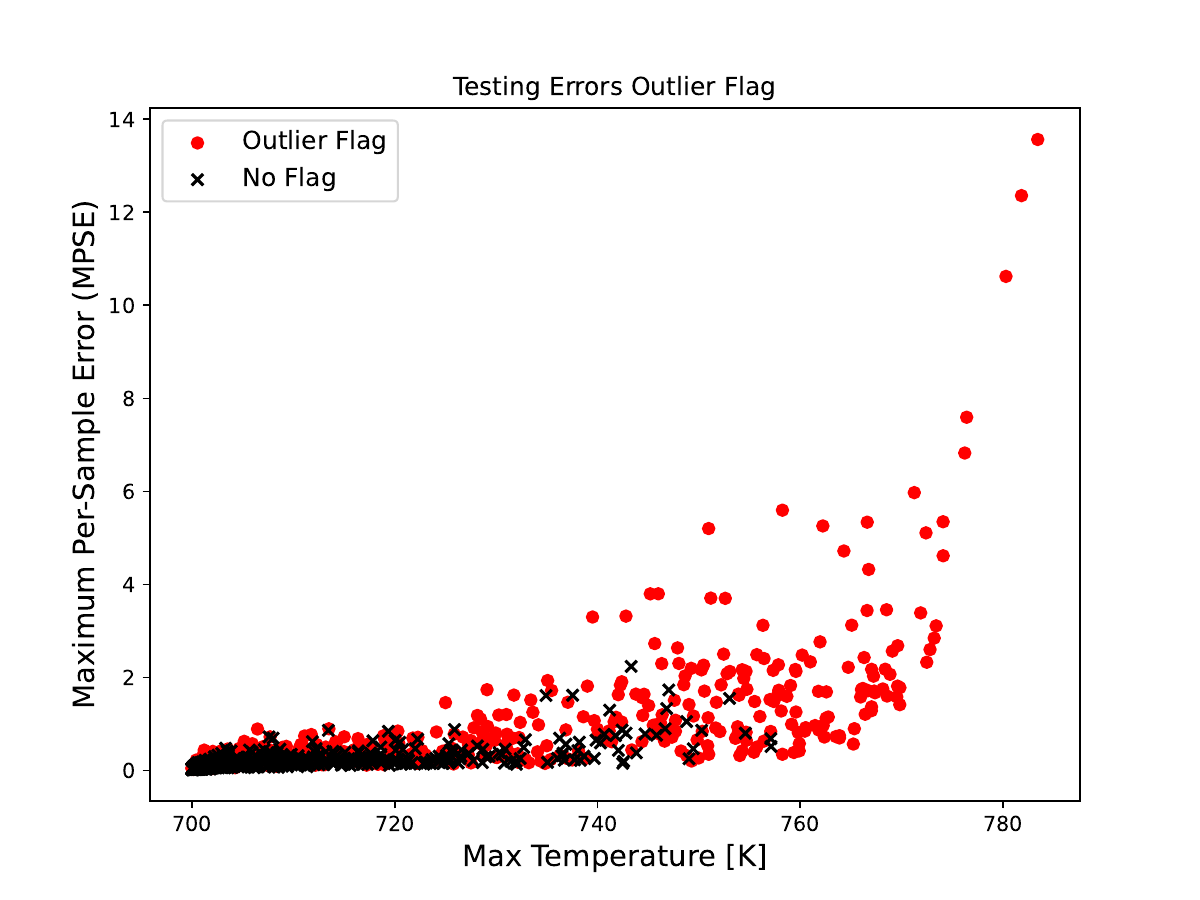}
		\caption{Sample outlier classification}
		\label{Scoure_1_classified}
	\end{subfigure}
	\hfill
	\begin{subfigure}[t]{0.48\textwidth}
		\centering
		\includegraphics[width=\textwidth]{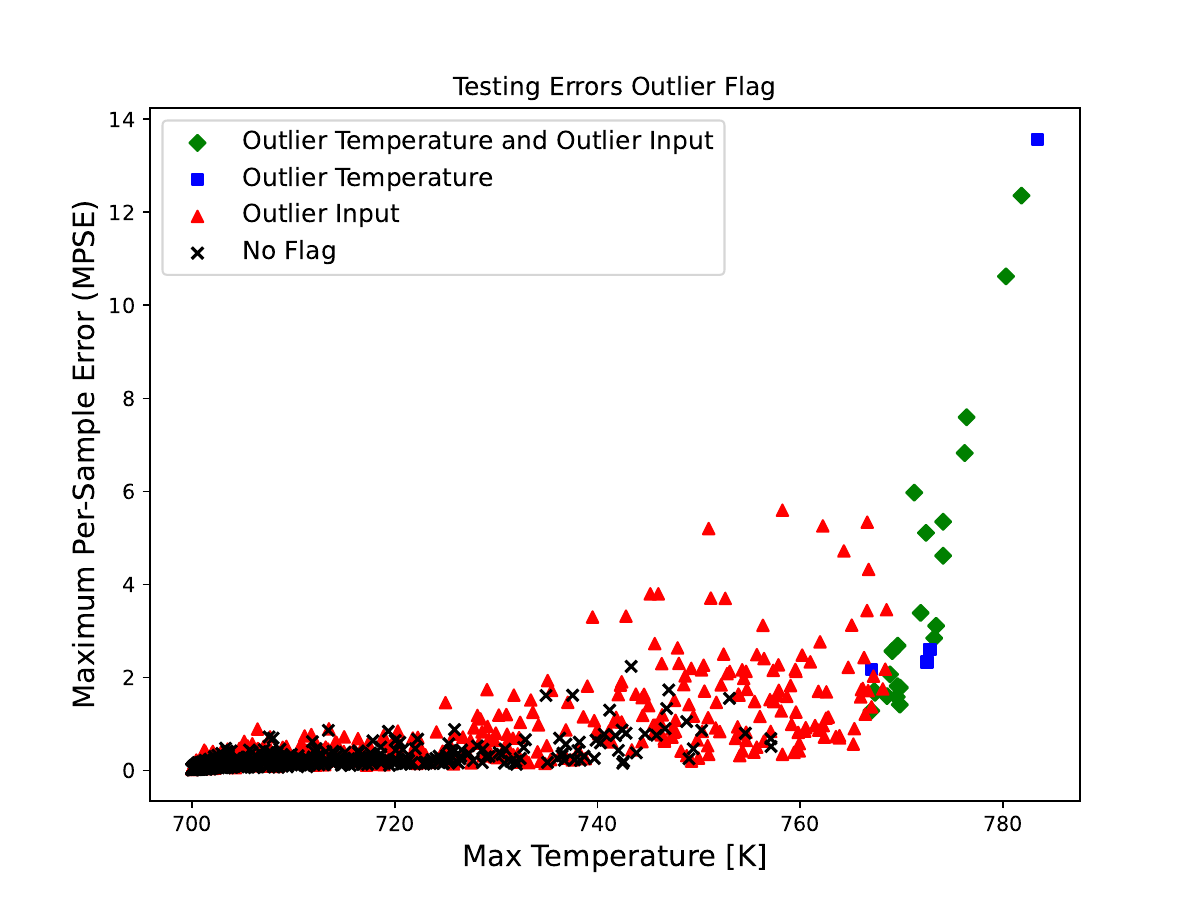} 
		\caption{Detailed classifications}
		\label{Scoure_1_detailed}
	\end{subfigure}
	\caption{\ktaddrem{Plots of the test data's maximum ground truth temperature versus MPSE for the PINN model trained on 400 samples from Source 1. The samples are demarcated through color coding based on if they are flagged as an outlier (circle) or not (x's) in (a) and further delineated by type of outlier as outlier temperature and outlier input (diamonds), outlier temperature (squares), outlier input only (triangles), and not an outlier (x's) in (b). While Figure (a) shows that the largest errors are associated with samples that are outliers relative to the training data, Figure (b) shows that the largest errors occur for outliers that can be identified from the largest predicted temperature fields. That is, input samples that are outliers (i.e., lie outside the convex hull of the training set) do not necessarily produce large errors.}}
	\label{Classified_testing_errors}
\end{figure}

In this subsection the PINN model is examined more closely to better understand \ktaddrem{the root cause for the error associated with each sample. Figure \ref{Classified_testing_errors} shows the MPSE for the PINN model trained on Source 1, color coded by whether or not each sample can be flagged as an outlier sample, defined as either:}
\ktaddrem{	
\begin{enumerate}
	\item \textbf{An Outlier Temperature}: A test sample whose predicted maximum temperature is greater that the largest predicted temperature of the training data set. 
	\item \textbf{An Outlier Input}: A source field sample which lies outside the convex hull of all the training input stochastic parameters. (for formulas to calculate the convex hull see \citeauthoryearkt{Preparata_2012}).
\end{enumerate}
}
\begin{figure}[ht]
	\centering
	\begin{subfigure}[t]{0.48\textwidth}
		\centering
		\includegraphics[width=\textwidth]{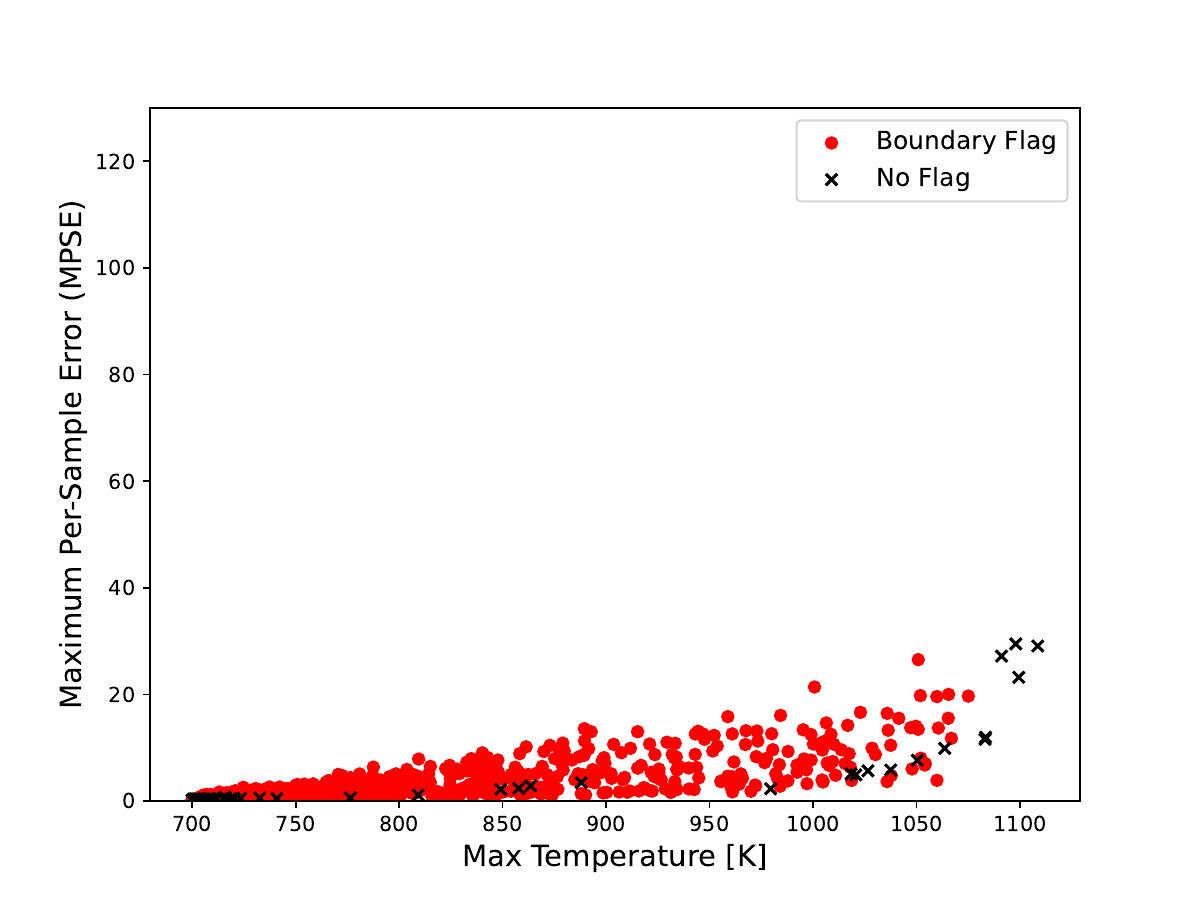}
		\caption{$w_b = 10^7$}
		\label{Bw_1e8}
	\end{subfigure}
	\hfill
	\begin{subfigure}[t]{0.48\textwidth}
		\centering
		\includegraphics[width=\textwidth]{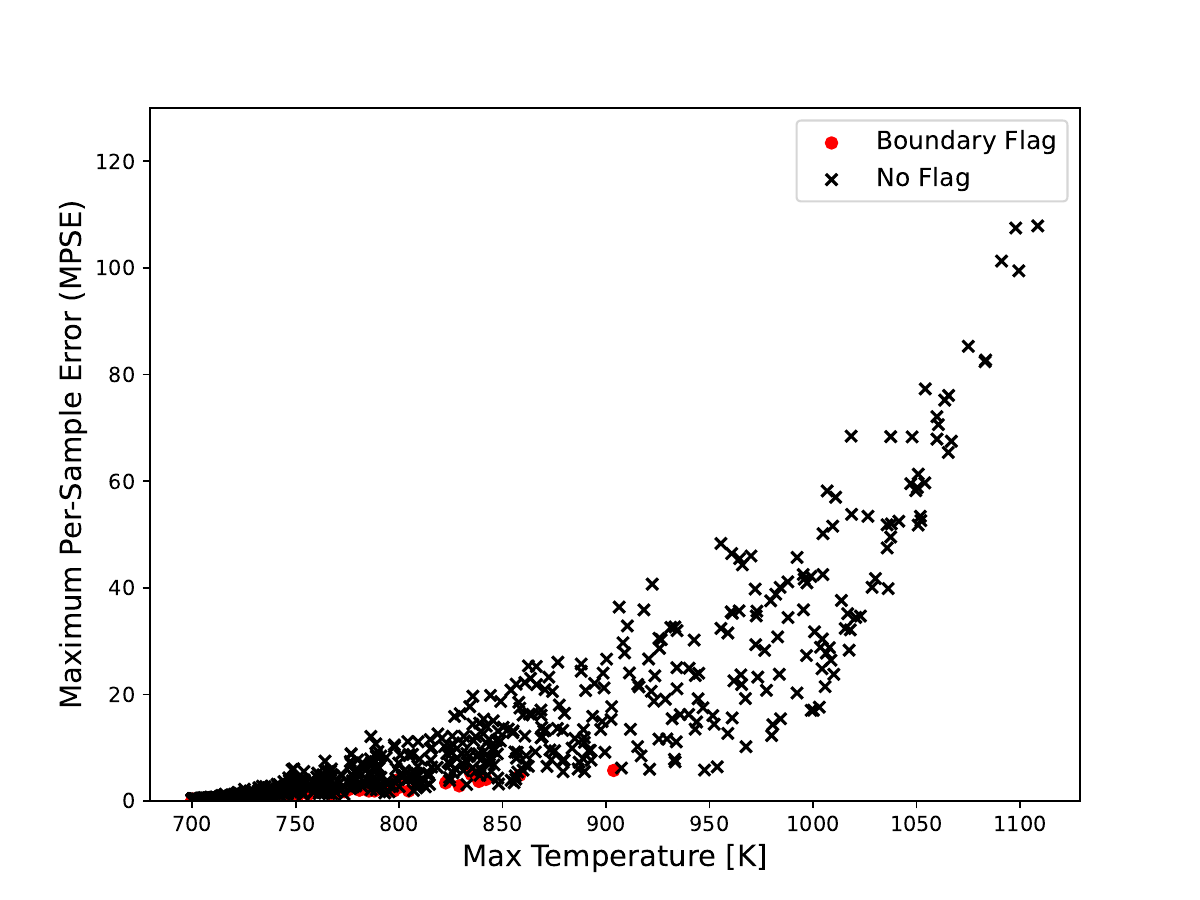}
		\caption{$w_b = 10^8$}
		\label{Bw_1e7}
	\end{subfigure}
	\caption{\ktaddrem{Plots of ground truth maximum temperature versus MPSE for Source 2 test data for the PINN model, coded by whether the MPSE is located on the boundary (circles) or interior of the domain (x's) for (a) $w_b=10^7$ and (b) $w_b=10^8$. Adjusting the boundary weight dramatically changes the MPSE. When the boundary weight ($w_b$) is set too high, training is over-constrained and the model fails to adequately learn to predict the interior of the domain. Relaxing this parameter allows training to more equitably distribute learning to the interior of the domain, lowering the overall error.}} 
	\label{B_w_comp}
\end{figure}

\noindent  \ktaddrem{If a sample meets either of these two definitions it can be classified as an outlier relative to the training data. The important conclusion here is that while extrapolation of samples is the source of the largest errors, it is not necessarily the case that extrapolated solutions based on input sample outliers cause large errors. However, samples associated with extrapolated response quantities of interest of the predicted solution field are strongly correlated with the largest errors. Fortunately, the flagging of these types of outliers can be performed simply by forward model execution of the trained neural network surrogate model for randomly selected samples of the source field. Predicted solution fields that have response quantities of interest that exceed those predicted during the training data can be flagged as potential outliers. This analysis can be intermittently conducted to identify new samples of full order models to simulate within an adaptive data augmentation methodology for constructing targeted training data in a more efficient way than random sampling, akin to the techniques discussed in Refs. (\citeauthoryearkt{Pickering_2022}, \citeauthoryearkt{Song_2024}).}

\ktaddrem{It is also worthwhile to evaluate the effect of the boundary term loss coefficient $w_b$ (i.e., see Eq. \eqref{Total_loss}) on whether the MPSE occurs in the interior or boundary of the domain. Based on the value of $w_b$ chosen for the comparative analysis, the models learn over the interior of the domain better than the boundary. However, Figure \ref{B_w_comp} plots the MPSE of the PINN model versus the maximum ground truth temperature for Source 2 test data for (a) $10\times$ scaling of the boundary weight to $w_b=10^8$ versus (b) by the boundary weight used in the comparative analysis, $w_b=10^7$. The plots are coded with cirlces for when the MPSE occurs on the boundary of the domain and x's for when the MPSE occurs in the interior of the domain. The model trained shown in Figure \ref{Bw_1e7} has significantly more error on the boundary ($88.1 \%$ of samples) than the interior of the domain ($11.9 \%$ of samples). On the other hand, attempting to compensate for this error discrepancy by increasing the boundary weight comes at the price of increasing the overall error, both in terms of $\ell^2$-norm and MPSE. Therefore, it must be concluded that the boundary weight, $w_b$, is a hyperparameter that strongly affects the training and test performance of the networks and must be adjusted as part of a global optimization. While the discussion above pertains to the PINN model, the same examination of outlier samples and boundary error samples in both the DDN and DeepONet models yield identical conclusions.}

\section{Conclusion} \label{Conclusions}
This study evaluates the training and testing error for neural network surrogate models \ktaddrem{for uncertainty propagation. While this study fits within the larger context of uncertainty quantification, the emphasis is placed on evaluating the sample-to-sample accuracy of the surrogate models with respect to ground truth full order models over the entire sample space. 
As prediction errors magnify at the tails of the probability distribution, model error is evaluated according to higher order norms, with the $\ell^\infty$-norm primarily being that which is reported.}  Three models are evaluated: (1) a fully connected feed forward network trained only on finite element data (DDN), (2) a fully connected feed forward network trained using a physics informed loss based on the weak form residual (PINN), and (3) a deep operator network architecture trained using a physics informed loss based on the weak form residual (PCA-DeepONet).   The heat conduction equation with a highly stochastic source term serves as the model problem for which this comparison is conducted. 

While all three models achieve low training error, the PINN model consistently outperforms both the DDN and PCA-DeepONet models when predicting the solution field on the test data in terms of maximum error (ME) and maximum per-sample errors (MPSEs). It is worth noting that all models have very low $\ell^2$-norm errors, consistent with other studies presented in literature; the errors in the tails of the distribution are less commonly investigated. \ktaddrem{ For the present study, all significant model errors are associated with cases pertaining to extrapolating beyond the training data. However, it is not necessarily the case that input samples outside the convex hull of the training set produce large errors. On the other hand, the solution field of model predictions whose response quantities of interest lie beyond that of the predicted solution fields of the training data consistently produce large errors. For example, the extrapolated sample with the worst prediction error for the PINN model is 15.6 times larger than its mean prediction error. In this study, an adaptive weighting strategy aimed toward increasing the contribution to the loss of extreme samples gives mixed results. The weighting strategy causes an overfitting, leading to a decrease in training error but an increase in testing error. Further, it may not be possible to find an optimal weighting function that is generalizable across different types of distributions. Instead, future efforts should explore the use of sample outlier identification as a method for data augmentation during the training process as an approach for efficiently constructing training data sets rather than naively expanding the convex hull of the input sample space through random sampling. Further, the establishment of model prediction error bounds becomes more imperative for such problems and should be the subject of future research.}

\section*{Data Availability Statement}
\noindent \ktaddrem{Some or all data, models, or code generated or used during the study are available in a repository online in accordance with funder data retention policies (\citeauthoryearkt{Wade_UQPINN}.)}

\section*{Acknowledgments}
\noindent Funding for this project was provided by the Office of Naval Research through the Naval Research Laboratory's Basic Research Program (No. N0001418WX00093). This work was supported in part by high-performance computer time and resources from the DoD High Performance Computing Modernization Program.
\def\url#1{}
\bibliography{referencesJEM}

\end{document}